\documentclass[lettersize,journal]{IEEEtran}
\usepackage{amsmath,amsfonts}
\usepackage{algorithmic}
\usepackage{algorithm}
\usepackage{array}
\usepackage[caption=false,font=normalsize,labelfont=sf,textfont=sf]{subfig}
\usepackage{textcomp}
\usepackage{stfloats}
\usepackage{url}
\usepackage{verbatim}
\usepackage{graphicx}
\usepackage{amssymb}
\usepackage{multirow}
\usepackage{vcell}
\usepackage{cite}
\usepackage{lineno}
\usepackage{hyperref}
\begin{document}

\title{Quaternion-valued Correlation Learning for Few-Shot Semantic Segmentation}

\author{Zewen Zheng,
Guoheng Huang$^{\ast}$,
Xiaochen Yuan$^{\ast}$,
Chi-Man Pun$^{\ast}$,
Hongrui Liu,
and Wing-Kuen Ling,
}

\maketitle

\begin{abstract}
Few-shot segmentation (FSS) aims to segment unseen classes given only a few annotated samples. Encouraging progress has been made for FSS by leveraging semantic features learned from base classes with sufficient training samples to represent novel classes. The correlation-based methods lack the ability to consider interaction of the two subspace matching scores due to the inherent nature of the real-valued 2D convolutions. In this paper, we introduce a quaternion perspective on correlation learning and propose a novel Quaternion-valued Correlation Learning Network (QCLNet), with the aim to alleviate the computational burden of high-dimensional correlation tensor and explore internal latent interaction between query and support images by leveraging operations defined by the established quaternion algebra. Specifically, our QCLNet is formulated as a hyper-complex valued network and represents correlation tensors in the quaternion domain, which uses quaternion-valued convolution to explore the external relations of query subspace when considering the hidden relationship of the support sub-dimension in the quaternion space. Extensive experiments on the PASCAL-5$^{i}$ and COCO-20$^{i}$ datasets demonstrate that our method outperforms the existing state-of-the-art methods effectively. Our code is available at \url{https://github.com/zwzheng98/QCLNet} and our article "Quaternion-valued Correlation Learning for Few-Shot Semantic Segmentation" was published in IEEE Transactions on Circuits and Systems for Video Technology, vol. 33,no.5,pp.2102-2115,May 2023,doi: 10.1109/TCSVT.2022.3223150.

\end{abstract}


\section{Introduction}\label{1.1}
\IEEEPARstart{N}{eural} network architectures such as Convolutional Neural Networks (CNNs) have made unprecedented progress in semantic segmentation. However, strong semantic segmentation models\cite{RN2,RN3} rely heavily on large-scale datasets with dense annotation, and models trained on such datasets often fail to handle novel object categories. As a promising direction, few-shot segmentation (FSS) is proposed to tackle the above challenge. It aims to train a model on a dataset with sufficient data and quickly adapt to the segmentation prediction of novel classes by using only a few annotations. Specifically, models are episodically trained on base classes with sufficient data samples and then located the target objects on novel classes based on the semantic information provided by the support set.

Fueled by the success of few-shot classification\cite{RN33,RN34}, current  FSS  models\cite{RN21,RN24,RN9,RN8,RN22,RN6}  often  use  a  metric-learning based framework, which utilizes the prototypes calculated from the support features to guide the query branch for semantic segmentation. However, performing metric learning on the base dataset with abundant annotated samples inevitably introduces a bias towards the seen classes rather than being ideally class-agnostic. More specifically, this learning mechanism easily forces the model to ‘remember’ objects outside the base class as negative samples and the embeddings of latent novel classes are over-smoothed. 

\begin{figure}[t]
\centering
\includegraphics[width=8.8 cm]{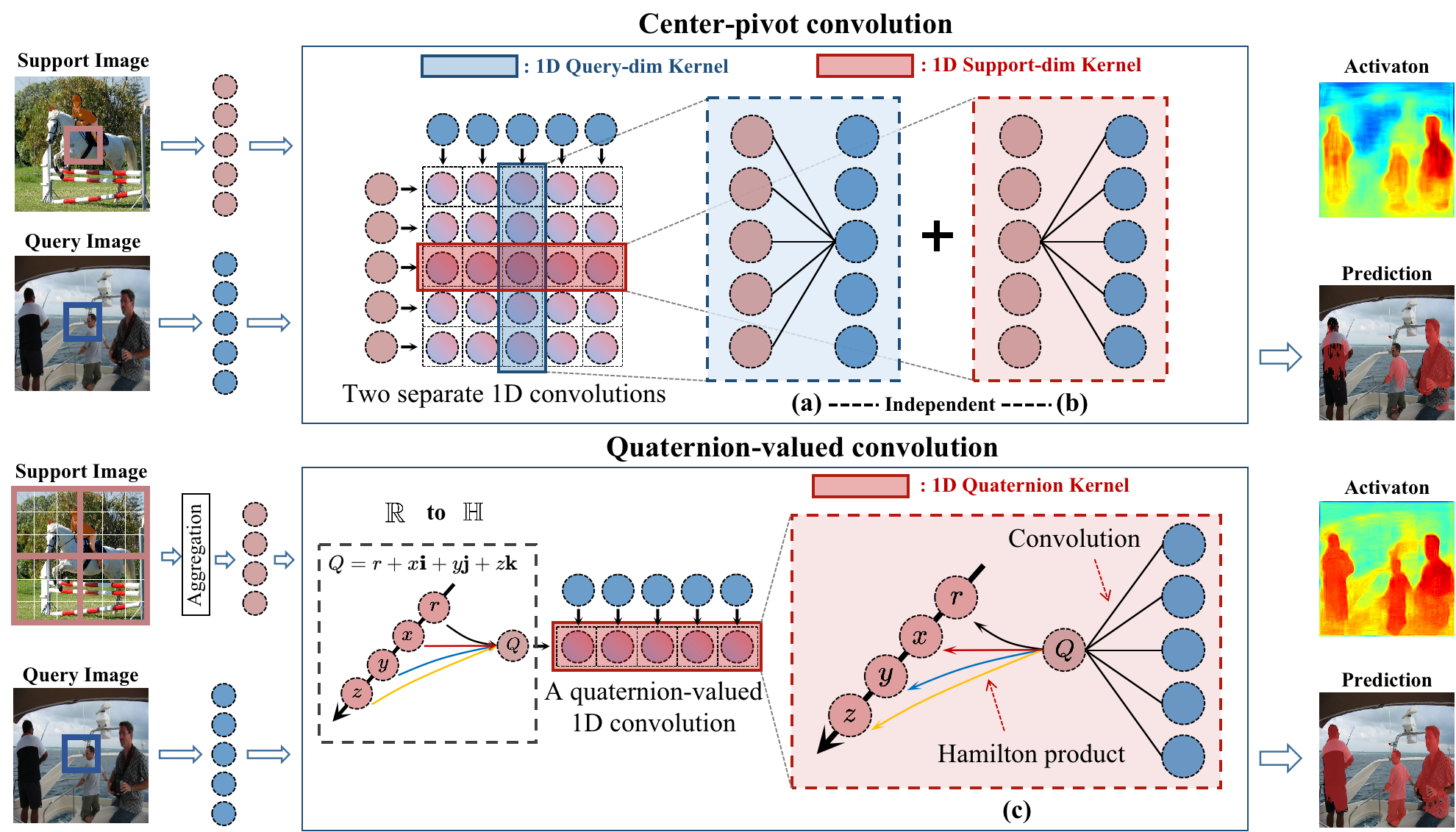}
\caption{Illustration of the difference between Center-pivot convolution\cite{RN38, RN85} and Quaternion-valued convolution in correlation learning. For ease of visualization, we show the 2D correlation for matching pixels across a query and support image. $\mathbb{R}$ to $\mathbb{H}$ means \textbf{R}eal-value to \textbf{H}yper-complex valued. Each black wire that connects two different pixel locations represents a single weight of the convolutional kernel. To efficiently filter the 2D correlation, the center-pivot convolution decomposees the 4D correlation learning into two independent subspace learning. Specifically, it factors the $5\times 5$ filter into support \textbf{(a)} and query \textbf{(b)} sub-dimension convolution kernels, which perform two different convolutions on separate 2D subspaces and thus cannot consider the interaction of the two subspace matching scores. In contrast, with the aggregation and encapsulation of support subspace of correlation maps into quaternion space (i.e., hypercomplex numbers), our method enhanced interaction between two subspaces by performing a double learning: \textbf{(1)} the convolution operator learns external/global relations among the elements of the query spatial dimension, \textbf{(2)} while the Hamilton product accomplishes the learning of the support subspace \textbf{(c)}.
\label{Figure1}}
\end{figure}

Such a problem can be easily solved by correlation learning methods in semantic correspondence task, which aims to construct the pixel-wise correlation between semantically similar images and exploring their internal latent relationships. Therefore, reformulating the FSS task as a semantic correlation learning problem can help the FSS model to capture more generic patterns, thereby improving the generalization. Recent works in correlation learning\cite{RN19, RN38} utilize  high-dimensional convolutions to aggregate correlation tensors and show significant efficiency in learning accurate relations. However, there are several challenges in  applying this real-valued high-dimensional convolution in FSS. First, the direct use of high-dimensional 4D convolution\cite{RN19,RN42} requires a large number of parameters, which increases the computational burden and is contrary to the original intention of lightweight design in FSS to ensure the generalization ability of the model. Second, there have been some attempts in utilizing center-pivot 4D convolution \cite{RN38, RN85} to reduce the amount of high-dimensional convolution parameters and achieve good performance. Despite their success, we notice that the design of this method still presents problems. Specifically, they factor the 4D correlation learning into two independent 2D subspace learning and simply employ two different 2D real-valued convolutions. Figure~\ref{Figure1} visualizes this factorization, which cannot consider the interaction of the two subspace matching scores. 

In light of this, we argue that an ideal model of correlation learning should be structured as a union dual-space learning process and be able to maintain the internal dependencies within the two subspaces. Therefore, real-valued convolution is not suitable for correlation learning due to its inherent nature of only dealing with a single feature space (i.e., spatial and channel relations). Hyper-complex valued convolutional neural networks solved this problem by introducing multidimensional algebra to CNN. As in Quaternion Convolutional Neural Networks (QCNN), by encapsulating the extra space information to the quaternion space, the Hamilton product allows QCNN to encode both internal relations that exist inside quaternion space and global relations of outside feature space at the same time.

In this paper, we propose a Quaternion-valued Correlation Learning Network (QCLNet). We move beyond real-valued space to explore the properties of quaternion algebra, e.g., \textit{Hamilton product}. As illustrated in Figure~\ref{Figure1}, by combining with the convolution operator, it allows the processing of support subspace of the correlation as a unique quaternion to perform a \textbf{double learning}: \textbf{(1)} the convolution operator learns external/global relations among the elements of the query spatial dimension, \textbf{(2)} while the Hamilton product accomplishes the learning of the support subspace. Specifically, to overcome the limitations of the computational burden and encapsulate the support information to the quaternion space, we propose a Correlation Aggregation Module (CAM) and utilize it to aggregate sparse information in high-dimensional correlation tensor. After that, we encapsulate the support spatial subspace of correlation maps in quaternion space (i.e., hypercomplex numbers) and propose a Quaternion Correlation Learning Module (QCLM) that consists of a series of quaternion-valued convolution and quaternion normalization (QN) to explore the external relations among the elements of the query spatial sub-dimension when considering the hidden relationship of the support subspace in the quaternion space. With correlation learning in the quaternion domain, the internal (i.e., the relations that exist inside support set) and external relations (i.e., edges or shapes features in query set) are learned simultaneously and thus the interactions of matching scores between query and support set are fully explored. As the interactions between the support and query set have been extracted, we further propose the Episodic Readout Module (ERM), which transforms quaternion features into real-valued features and utilizes low-level query features to refine the segmentation results. The contributions of this work are summarized as follows:
\begin{itemize}
\item We propose the Quaternion-valued Correlation Learning Network (QCLNet), which explicitly explores interactions of matching scores between query and support images by leveraging operations defined by the established quaternion algebra.

\item We introduce a quaternion perspective on correlation learning and propose a novel quaternion correlation learning module, QCLM, which encapsulates the support sub-dimension in quaternion space and performs quaternion-valued convolution with our proposed theoretically correct quaternion normalization (QN) to explore the interaction of two subspaces of the correlation.

\item We propose a correlation aggregation module (CAM) and episodic readout module (ERM) to aggregate sparse information of the correlation tensors and adaptively refine the segmentation results with the low-level query features.

\item Our method achieves better performance and effectiveness than other state-of-the-art methods on two FSS benchmark datasets: PASCAL-5$^{i}$ and COCO-20$^{i}$.

\end{itemize}

\section{Related Work}
\subsection{Semantic Segmentation}
Semantic segmentation is a fundamental and challenging task that has gained interest in the computer vision community for decades due to its ability to provide pixel-wise dense semantic prediction\cite{RN43, RN44, RN2, RN46}. Since the great success of fully convolutional neural networks in the field of semantic segmentation, various networks , such as Deeplab\cite{RN48}, PSPNet\cite{RN3}, UNet\cite{RN2} and SegNet\cite{RN43} have been proposed in this field. Contextual information provides surrounding hints to help identify individual elements, thus later works contribute many benchmark blocks, such as the pyramid pooling module\cite{RN3}, deformable convolution\cite{RN53}, non-local module\cite{RN59} to help enlarge the receptive field of the model and achieved good performance. However, powerful segmentation models cannot be extended to unseen class segmentation scenarios without updating the parameters of the model.
\subsection{Few-Shot Segmentation}
Few-Shot Segmentation (FSS) requires the model to quickly segment the target region in the input image with only a few annotated samples. Almost all existing models use two-branch architecture design to implement meta-learning. OSLSM\cite{RN20} is the pioneering work for FSS, which includes two branch structures: conditional branch and segmentation branch. The conditional branch is used to generate classifier weights for the query image of each task. Afterward, global or multiple prototype-based methods were designed under these two-branch paradigm, representative models include PANet\cite{RN22}, PMMs\cite{RN8}, CANet\cite{RN24}, PPNet\cite{RN7} and PFENet\cite{RN6}. PANet\cite{RN22} uses prototype alignment regularization to provide high-quality prototypes that are representative of each semantic class. The work of \cite{RN64} generalizes FSS to a multi-class task and mainly studies the application of incremental learning in few-shot tasks. However, as methods based on prototypes have apparent limitations, e.g., performing metric learning on the base dataset inevitably introduces a bias towards the seen classes rather than being ideally class-agnostic. Recent works\cite{RN85} attempted to utilize efficient 4D convolution consisting of two decoupled 2D convolutions to fully exploit the multi-level correlations. However, this correlation-based methods still face the challenge of lacking the ability to simultaneously consider the internal interaction of the two subspaces due to the inherent nature of the real-valued 2D convolutions.
\subsection{Semantic Correspondences}
In recent years, finding dense semantic correspondences has been studied extensively in low-level vision. The objective of semantic correspondence is to find reliable correspondences between a pair of images with challenges of large intra-class variations\cite{RN38, RN42, RN40,RN19}. This setting is very similar to few-shot semantic segmentation, which aims to use the semantic features of the support set to guide the query branch for semantic segmentation. Rocco \textit{et al}. \cite{RN19} introduce the neighborhood consensus network that uses 4D convolution to learn local geometric constraints between neighboring correspondences, and thus  requires a large number of parameters. Following the work, recent methods \cite{RN32, RN42} also adopt 4D convolution in a similar manner. The work of \cite{RN38} resolves the former problem (quadratic complexity) by separating a 4D convolution into two center-pivot 2D kernels and downsampling the 4D cost-volume to maintain small memory footprints. Despite the center-pivot 4D convolution reducing the computational burden caused by the use of high-dimensional convolution, the hidden internal relations of matching scores are unfortunately unexplored. In our work, we use quaternion neural networks to fully explore the interactions of matching scores while yielding substantial improvements in parameter size.
\subsection{Complex and Quaternion Networks}
In various deep learning application areas\cite{RN60,RN61,RN62}, such as images, 3D audio, multi-sensor signals or human-pose estimation, some efforts have been made to extend real-valued neural networks to other number fields. Complex-valued neural networks\cite{RN63} or quaternion neural networks \cite{RN60, RN26, RN27, RN28} (QNN) have been proposed to encapsulate multidimensional input features. In \cite{RN26}, a deep quaternion network is proposed, which simply replaces the real multiplications with quaternion ones. The work of \cite{RN27, RN29} further explores the application of QNN and QCNN to image processing, where they use Hamilton product to embed the three components (R,G,B) of a given pixel in a quaternion and maintain its internal dependencies in the subsequent convolution process. Similarly, we believe that few-shot correlation learning should be constructed as a dual-space learning process since it needs to consider both support and query subspace information. Therefore, instead of introducing multi-dimensional algebra to maintain the structural dependence of the three components (R,G,B), our approach mainly aims to achieve correlation learning by introducing QCNN to explore the external relations of query subspace when considering the hidden relationship of the support sub-dimension in the quaternion space.

\begin{figure*}
\centering
\includegraphics[width=18.0 cm]{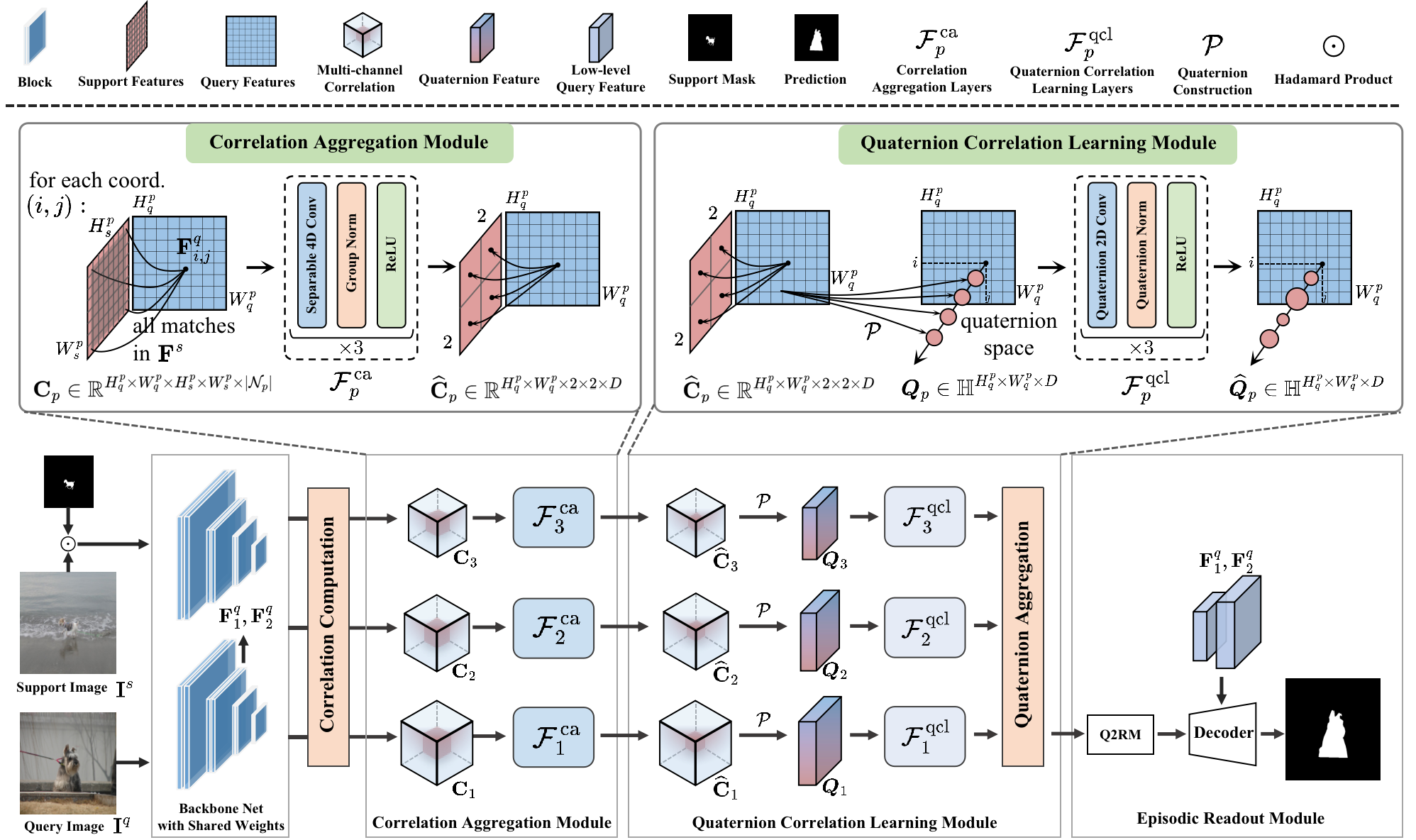}
\caption{Overview of our Quaternion-valued Correlation Learning Network (QCLNet), which consists of correlation aggregation module, quaternion correlation learning module and episodic readout module. Form the query and support feature maps $\mathbf{F}^{q}$, $\mathbf{F}^{s}$, their matching scores are computed and stored in the 4D correlation tensor $\mathbf{C}_{p}$. Then the support spatial dimensions of correlation tensor are gradually reduced by correlation aggregation layers. In order to use the operations defined by the established quaternion algebra for correlation learning, we transform the encoded correlation $\widehat{\mathbf{C}}_{p}$ into quaternion features $\boldsymbol{Q}_{p}$ and further use the quaternion correlation learning layers to produce the output quaternion feature $\widehat{\boldsymbol{Q}}_{p}$.
\label{Figure2}}
\end{figure*}

\begin{table}
\renewcommand\arraystretch{1.3}
\centering
\caption{The definition of notations.}
\label{tab1}
\begin{tabular}{ll} 
\hline
Notations                                                   & Description                                               \\ 
\hline
$\mathcal{D}_{\text {base }}, \mathcal{D}_{\text {novel }}$ & Base data, novel data                                     \\
$\mathbf{F}^s, \mathbf{F}^q$                                & Support and query feature~given in (7)                    \\
$\mathbf{C}_p$                                              & Correlation maps given in (9)                             \\
$\mathbf{k}$                                                & Separable~4D convolution kernel~                          \\
$\mathbf{u}, \mathbf{x}$                                    & Query and support 2D spatial coordinates                  \\
$\Psi\left(\cdot\right)$                                    & A set of~neighborhood coordinates centered on u and x     \\
$\mathcal{F}^\text {ca}_\text {p}\left(\cdot\right)$        & Correlation aggregation module~defined in (11)            \\
$\boldsymbol{Q}_p$                                          & Quaternion feature~given in (12)                          \\
$\mathbf{W}_q$                                              & Quaternion convolution weights                            \\
$(\cdot)^\mathrm{H}$                                        & Conjugate transpose operator                              \\
$\tilde{\mathbf{C}}_\mathbf{q q}$                           & Augmented covariance matrix~given in (16)                 \\
$\mathrm{QN}(\cdot)$                                        & Quaternion normalization~defined in (18)                  \\
$\boldsymbol{\mu}_q, \boldsymbol{\sigma}^2$                 & Quaternion mean~given in (19) and variance~given in (20)  \\
$\mathcal{F}^\text {qcl}_\text {p}\left(\cdot\right)$       & Quaternion convolutional block~defined in (21)            \\
$\mathrm{up}_[\times 2](\cdot)$                             & Upsampling 2x operator                                    \\
$\mathbf{F}_r$                                              & Real-valued feature~given in (23)                         \\
$\mathcal{G}(\cdot)$                                        & Global average pooling                                    \\
$\mathcal{P}_i(\cdot)$                                      & Linear projection                                         \\
$\tilde{\mathbf{M}}_{q}$             
    & Predicted mask                                        \\
\hline
\end{tabular}
\end{table}

\section{Quaternion Algebra}
This section introduces the necessary background of quaternion for this paper. Quaternion is a kind of hypercomplex number of rank 4, being a direct non-commutative extension of complex-valued numbers. Along with Hamilton products, quaternion algebra forms the crux of our proposed approaches.
\textbf{Quaternion} A quaternion Q in the quaternion domain $\mathbb{H}$, i.e., $Q \in \mathbb{H}$, can be represented as:  
\begin{equation}
Q=r+x \mathbf{i}+y \mathbf{j}+z \mathbf{k},\label{1}
\end{equation}

\noindent where $r$, $x$, $y$, and $z$ are real numbers, and $\mathbf{i}$, $\mathbf{j}$, and $\mathbf{k}$ are the quaternion unit basis. In a quaternion, $r$ is the real part, while $x\mathbf{i} + y\mathbf{j} + z\mathbf{k}$ with $\mathbf{i}^2 = \mathbf{j}^2 = \mathbf{k}^2 = \mathbf{ijk} = -1$ is the imaginary part. A pure quaternion is a quaternion whose real part is 0,  resulting in the vector $Q=x \mathbf{i}+y \mathbf{j}+z \mathbf{k}$. Operations on Quaternions are defined in the following.

\noindent \textbf{Addition} The addition of two Quaternions is defined as:

\begin{equation}
\begin{array}{r}
Q+P=Q_{r}+P_{r}+\left(Q_{x}+P_{x}\right) \mathbf{i} \\
\quad+\left(Q_{y}+P_{y}\right) \mathbf{j}+\left(Q_{z}+P_{z}\right) \mathbf{k},
\end{array}\label{2}
\end{equation}

\noindent where $Q$ and $P$ with subscripts denote the real value and imaginary components of Quaternion $Q$ and $P$.

\noindent\textbf{Scalar Multiplication} The Multiplication with scalar $\alpha$ is defined as:

\begin{equation}
\alpha Q=\alpha r+\alpha x \mathbf{i}+\alpha y \mathbf{j}+\alpha z \mathbf{k},\label{3}
\end{equation}

\noindent\textbf{Conjugate} The conjugate $Q^{*}$ of $Q$ is defined as:
\begin{equation}
Q^{*}=r-x \mathbf{i}-y \mathbf{j}-z \mathbf{k},\label{4}
\end{equation}

\noindent\textbf{Norm} The unit Quaternion $Q^{\triangleleft}$ is defined as:
\begin{equation}
Q^{\triangleleft}=\frac{Q}{\sqrt{r^{2}+x^{2}+y^{2}+z^{2}}},\label{5}
\end{equation}

\noindent\textbf{Hamilton Product}
The Hamilton product is used to replace the standard real-valued dot product, which represents the multiplication of two quaternions $Q$ and $P$. It is defined as:
\begin{equation}
\begin{aligned}
Q \otimes P &=\left(Q_{r} P_{r}-Q_{x} P_{x}-Q_{y} P_{y}-Q_{z} P_{z}\right) \\
&+\left(Q_{x} P_{r}+Q_{r} P_{x}-Q_{z} P_{y}+Q_{y} P_{z}\right) \mathbf{i} \\
&+\left(Q_{y} P_{r}+Q_{z} P_{x}+Q_{r} P_{y}-Q_{x} P_{z}\right) \mathbf{j} \\
&+\left(Q_{z} P_{r}-Q_{y} P_{x}+Q_{x} P_{y}+Q_{r} P_{z}\right) \mathbf{k},
\end{aligned}\label{6}
\end{equation}
\noindent which intuitively encourages inter-latent interaction between quaternion $Q$ and $P$. Therefore, hamilton product plays a crucial role in quaternion neural networks. As illustrated in Figure~\ref{Figure4}, the quaternion-weight components are shared through multiple quaternion-input parts during the Hamilton product, exploring hidden relations within elements. In this work, we use Hamilton product extensively for correlation learning, which is at the heart of the better interaction ability of FSS.

\section{Method}
We adopt the meta-learning setting to conduct FSS. Typically, in FSS, given two disjoint image sets $\mathcal{D}_{\text {base }}$ and $\mathcal{D}_{\text {novel }}$ ($\mathcal{D}_{\text {base }} \cap \mathcal{D}_{\text {novel }}=\varnothing$), models are required to learn the correlation interaction on $\mathcal{D}_{\text {base }}$ with sufficient data and test on $\mathcal{D}_{\text {novel }}$. Both $\mathcal{D}_{\text {base }}$ and $\mathcal{D}_{\text {novel }}$ contain several episodes, and each of them is formed by a support set $S=\left(\mathbf{I}_{i}^{s}, \mathbf{M}_{i}^{s}\right)_{i=1}^{K}$ and a query set $Q=\left(\mathbf{I}^{q}, \mathbf{M}^{q}\right)$ of the same class, where $K, \mathbf{I}_{i}^{s}, \mathbf{M}_{i}^{s}, \mathbf{I}^{q}$ and $ \mathbf{M}^{q}$ represent the number of shot, the support image, the support binary mask, the query image and the query binary mask respectively. During the training of FSS, the model is optimized to segment the objects in the query image $ \mathbf{I}^{q}$ by taking $S$ and  $ \mathbf{I}^{q}$  in each episode $(S,Q)$ as inputs. Segmentation performance is evaluated on $\mathcal{D}_{\text {novel }}$ across all the test episodes. 

As most FSS models \cite{RN8,RN24,RN7,RN6,RN22}, the 1-way scenario is our focus in this paper, i.e., each pixel is classified as foreground or background. And we consider the 1-shot setting (i.e., $K=1\;in\;S$) to clearly illustrate our proposed approach.

\subsection{Overview of QCLNet}\label{4.2}
We propose a novel FSS framework, Quaternion-valued Correlation Learning Network (QCLNet), as shown in Figure~\ref{Figure2}, to explore internal latent interaction between query and support images by leveraging operations defined by the established quaternion algebra. In this section, we first briefly describe the multi-channel correlation computation in Section \ref{4.3}. In Section \ref{4.4}, to transform the high-dimensional correlation tensor to the quaternion space for correlation learning, we propose the correlation aggregation module (CAM) to effectively aggregate the local information of correlation to a global context. Then, in Section \ref{4.6}, the quaternion correlation learning module (QCLM) is used to simultaneously exploit the information of two sub-dimensions (i.e., the support and query) and consider their interaction. Finally, in Section \ref{4.7}, a readout module is used to fuse corresponding low-level query feature and refine the segmentation results. As such, QCLNet can be trained in an end-to-end manner and can transfer correlational knowledge from the seen to unseen domain (meta-testing). For convenience, a summary of notations is given in Table \ref{tab1}.

\subsection{Multi-channel Correlation Computation}\label{4.3}
Following the finding by\cite{RN6,RN24,RN18}, we fix the backbone weights, use a rich of features from the intermediate layers for multi-channel correlation computation. Specifically, for ResNet with layers divided into four groups (\textit{block1-4}), the spatial size of feature maps in each block is the same. Then we use the feature maps after \textit{block2} to produce a sequence of $\mathcal{N}$ pairs of intermediate feature maps $\left\{\left(\mathbf{F}^{q}_{i}, \mathbf{F}^{s}_{i}\right)\right\}_{i=1}^{\mathcal{N}}$. Denoting the above process as $\mathcal{B}$, given the support/query images $\mathbf{I}^s/\mathbf{I}^q$, we utilize support mask $\mathbf{M}^{s}$ to filter out the background area and obtain the intermediate feature maps:
\begin{equation}
\mathbf{F}_{i}^{s}=\mathcal{B}\left(\mathbf{I}^{s}\right) \odot \mathcal{R}_i(\mathbf{M}^{s}) , \quad \mathbf{F}_{i}^{q}=\mathcal{B}\left(\mathbf{I}^{q}\right),\label{7}
\end{equation}
where $\odot$ is element-wise multiplication, and $\mathcal{R}_i(\cdot)$ denotes a function that resizes $\mathbf{M}^{s}$ along the channel dimension. To obtain the multi-channel correlation $\mathbf{c}_{i}\left(\mathbf{x}^{\mathrm{q}}, \mathbf{x}^{\mathrm{s}}\right) \in \mathbb{R}^{H^i_q \times W^i_q \times H^i_s \times W^i_s}$, we compute the cosine similarity between query and masked support features such that: 
\begin{equation}
\mathbf{c}_{i}\left(\mathbf{x}^{\mathrm{q}}, \mathbf{x}^{\mathrm{s}}\right)=\operatorname{ReLU}\left(\frac{\mathbf{F}_{i}^{\mathrm{q}}\left(\mathbf{x}^{\mathrm{q}}\right) \cdot \mathbf{F}_{i}^{\mathrm{s}}\left(\mathbf{x}^{\mathrm{s}}\right)}{\left\|\mathbf{F}_{i}^{\mathrm{q}}\left(\mathbf{x}^{\mathrm{q}}\right)\right\|\left\|\mathbf{F}_{i}^{\mathrm{s}}\left(\mathbf{x}^{\mathrm{s}}\right)\right\|}\right), \label{8}
\end{equation}
where $\mathbf{x}^{\mathrm{s}},\mathbf{x}^{\mathrm{q}} \in \mathbb{R}^{2}$ is the pixel coordinate of feature maps $\mathbf{F}_{l}^{s}$ and $\mathbf{F}_{l}^{q}$ respectively. As done in \cite{RN18}, we record correlation maps computed from the intermediate features in the same blocks to form a multi-channel correlation: 
\begin{equation}
\mathbf{C}_{p}=\mathcal{F}_{\text {concat }}(\{\mathbf{c}_{i}\}_{i \in \mathcal{N}_{p}}) \in \mathbb{R}^{H^p_q \times W^p_q \times H^p_s \times W^p_s \times |\mathcal{N}_p|},\label{9}
\end{equation}
where $\mathcal{N}_{p}$ denotes the CNN layer indices belonging to the same block, $\mathcal{F}_{\text {concat }}(\cdot)$ concatenates the input intermediate features and considers $|\mathcal{N}_{p}|$ as the feature channel, $H^p_q$, $W^p_q$, $H^p_s$, $W^p_s$ represents the spatial resolution of the multi-channel correlation tensor. 

\subsection{Correlation Aggregation Module}\label{4.4}
\subsubsection{Motivation}
Due to the high-dimensional properties of the correlation $C_p$,  learning the internal interactions between support and query feature requires extremely large computation (quadratic complexity) . As an alternative, the work of \cite{RN42, RN85} stores only the most promising matching scores (i.e.,  top K or center values ) in $C_p$ to effectively reduce the spatially sparse information. However, such an approach would ignore the neighborhood information in $C_p$, which is proven to be extremely critical for correlation learning \cite{RN19}. 

To alleviate the above limitations, we gradually reduce and aggregate the spatial information in high-dimensional correlation tensors $C_p$ by controlling the different strides of the 2D convolution in the separable 4D convolution. After correlation aggregation, correlation tensor encapsulated into quaternion space is utilized for subsequent correlation learning. Details of CAM are as follows.

\subsubsection{Module Structure}
As shown in Figure~\ref{Figure2}, the CAM achieve correlation aggregation by applying separable 4D convolution, group normalization (GN)\cite{RN69}, ReLU activation, sequentially. In separable 4D convolution, we use two 2D convolution kernels to perform aggregation of two spatial dimensions (i.e., support and query) with different strides, where the support spatial dimension is reduced to $(2,2)$ and the query spatial dimension remains the same as $(H^p_q, W^p_q)$. Meanwhile, the separable 4D convolution also projects $C_p$ at separate 2D subspaces to embed the $|\mathcal{N}_p|$ to a fixed dimension $D$. We now show the factorization of the separable 4D convolution kernel $\mathbf{k}$ into two 2D spatial convolution kernel $\mathbf{k}_{s}$ and $\mathbf{k}_{q}$ :

\begin{small}
\begin{equation}
\begin{aligned}
(\mathbf{k}*\mathbf{c})(\mathbf{u}, \mathbf{x})&= \sum_{\mathbf{x}^{\prime} \in \Psi\left(\mathbf{x}\right)} \mathbf{k}_{s}(\mathbf{x}^{\prime}-\mathbf{x})  \left[
\sum_{\mathbf{u}^{\prime} \in \Psi(\mathbf{u})} \mathbf{k}_{q}(\mathbf{u}^{\prime}-\mathbf{u}) \mathbf{c}(\mathbf{u}^{\prime}, \mathbf{x}^{\prime}) \right] \\
&=\mathbf{k}_{s}\left(\mathbf{x}\right)* \left[\mathbf{k}_{q}(\mathbf{u}) * \mathbf{c}\left(\mathbf{u}, \mathbf{x}\right)\right],\label{10} \\
\end{aligned}
\end{equation}
\end{small}

\noindent where $u$ and $x$ are the query and support 2D spatial coordinates in correlation maps, and $\Psi\left(\cdot\right)$ denotes a set of neighborhood centered on 2D spatial coordinate $u$, $x$. Overall, the CAM is defined as:
\begin{equation}
\widehat{\mathbf{C}}_{p} = \mathcal{F}^{\text {ca}}_{\text {p}}(\mathbf{C}_{p})\in \mathbb{R}^{H^p_q \times W^p_q \times 2 \times 2 \times D}.\label{11}
\end{equation}

\subsection{Quaternion-valued Correlation Learning Module}\label{4.6}
\subsubsection{Motivation}
Existing correlation-based methods \cite{RN38, RN85} use center-pivot 4D convolutions to squeeze the matching scores of the hypercorrelation while reducing the large computational burden caused by high-dimensional correlation. However, since this method factors the 4D correlation learning into two independent 2D subspace learning, the interaction of the two subspace matching scores cannot be fully considered. Therefore, decomposing the union correlation learning into two independent real-valued convolutions is not ideal. 

Inspired by recent hyper-complex convolutional approaches \cite{RN27, RN29, RN60, RN61}, we consider introducing multidimensional algebra (i.e., quaternions algebra specifically) to few-shot correlation learning to alleviate the above problem. After the correlation aggregation, support spatial dimension in the correlation tensor is efficiently aggregated and becomes tractable. Then, by encapsulating the aggregated support subspace of the correlation maps into the quaternion space, Hamiltonian product in quaternion algebra allows quaternion convolution encodes both internal relations that exist inside quaternion space and global relations of outside feature space at the same time. 

\begin{figure}[t]
\centering
 \includegraphics[width=8.8 cm]{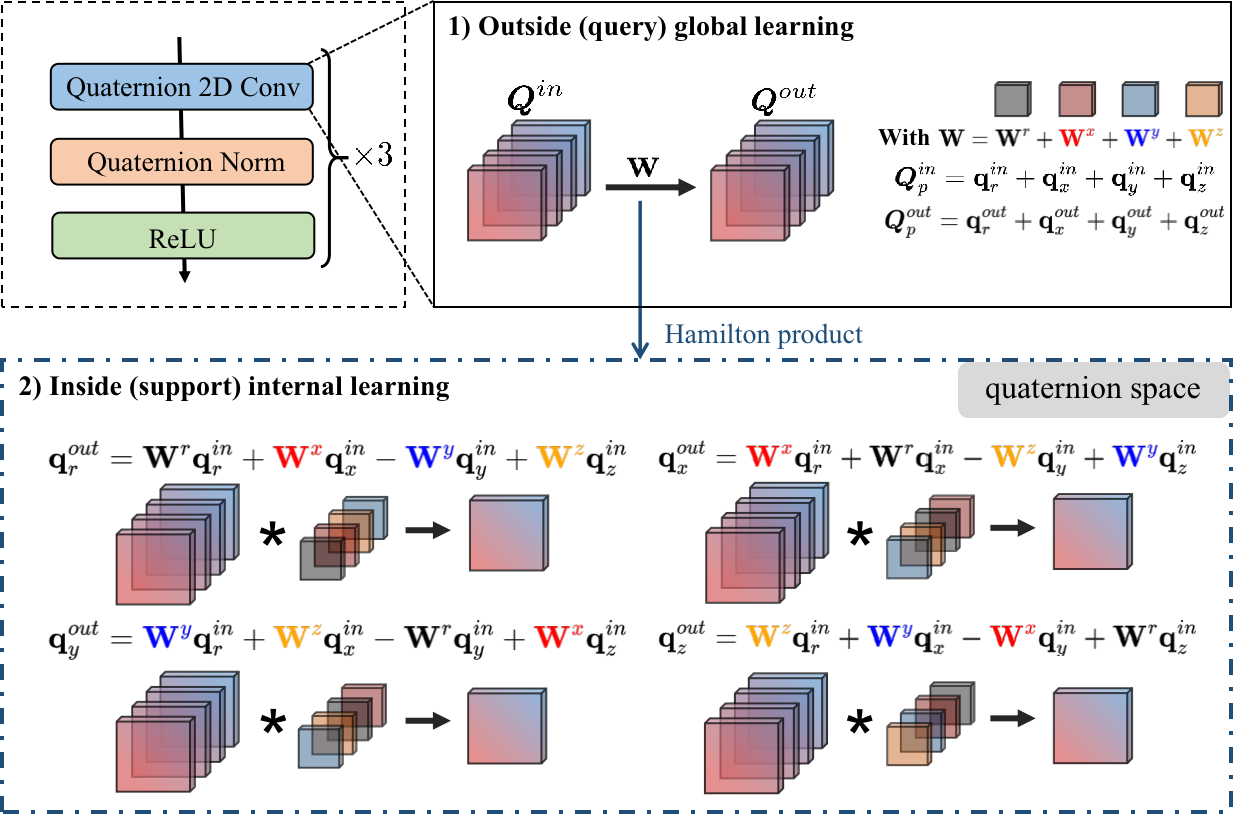}
\caption{Illustration of Quaternion-valued convolution.
\label{Figure3}}
\end{figure} 

\subsubsection{Quaternion-valued Correlation Learning}
After the correlation aggregation, each vector of support spatial dimension in the correlation tensor is efficiently aggregated and has larger receptive fields, i.e., $\mathbb{R}^{H^p_q \times W^p_q \times H^p_s \times W^p_s \times |\mathcal{N}_p|}\rightarrow \mathbb{R}^{H^p_q \times W^p_q \times 2 \times 2 \times D}$. To apply quaternion algebra to correlation learning, we propose a quaternion representation that preserves the support spatial dimension of the correlation maps $\widehat{\mathbf{C}}_{p}$  and encapsulates it as a quaternion-valued feature maps $\boldsymbol{Q}_{p} \in \mathbb{H}^{H_{q}^{p} \times W_{q}^{p}  \times D}$:

\begin{equation}
\begin{aligned}
\boldsymbol{Q}_{p} & =\  \widehat{\mathbf{C}}_{p}^{}(\mathbf{u}, \mathbf{x}_{(0,0)})+\widehat{\mathbf{C}}_{p}(\mathbf{u}, \mathbf{x}_{(0,1)})\mathbf{i}\\
& \  +\widehat{\mathbf{C}}_{p}(\mathbf{u}, \mathbf{x}_{(1,0)})\mathbf{j}+\widehat{\mathbf{C}}_{p}(\mathbf{u}, \mathbf{x}_{(1,1)})\mathbf{k},
\end{aligned}\label{12}
\end{equation}

\noindent where $\mathbf{u}$, $\mathbf{x}$ are the 2D spatial coordinate of query and support in the correlation maps $\widehat{\mathbf{C}}_{p}$. As illustrated in Figure~\ref{Figure2}, encapsulating support subspace in a quaternion allows treating each vector of query spatial dimension as a single entity and thus to preserving support intra-subspace relations. Therefore, we further utilize quaternion-valued 2D convolution to explore the external relations of query subspace when considering the hidden relationship of the support sub-dimension in the quaternion space. 
In order to define the convolutional operation in the quaternion domain, we first define the Standard 2D convolution process as:

\begin{equation}
\widehat{\mathbf{x}}=\phi\left(\mathbf{W}_{r} \otimes \mathbf{x}+\mathbf{b}_{\mathrm{r}}\right), \label{13}
\end{equation}
where $\mathbf{W}_{r} \otimes \mathbf{x}$ performs the convolution between the weight matrix $\mathbf{W}_{r}$ and the input $\mathbf{x}$ , $\mathbf{b}_{\mathrm{r}}$ is the bias and $\phi(\cdot)$ is any activation function. Then, we represent the quaternion weight matrix as $\mathbf{W}_q=\mathbf{W}_r+\mathbf{W}_x \mathbf{i}+\mathbf{W}_y \mathbf{j}+\mathbf{W}_z \mathbf{k}$,  the quaternion input as $\mathbf{x}=\mathbf{q}_{r}+\mathbf{q}_{x} \mathbf{i}+\mathbf{q}_{y} \mathbf{j}+\mathbf{q}_{z} \mathbf{k}$ and the quaternion bias as $\mathbf{b}_q=\mathbf{b}_{\mathrm{r}}+\mathbf{b}_{\mathrm{x}} \mathbf{i}+\mathbf{b}_{\mathrm{y}} \mathbf{j}+\mathbf{b}_{\mathrm{z}} \mathbf{k}$. Therefore, $\mathbf{W} \otimes  \mathbf{x}$ in Equation~\ref{13}, is performed by a vector multiplication between two quaternions, i.e., by the Hamilton product: 

\begin{small}
\begin{equation}
\begin{aligned}
\mathbf{W}_q \otimes \mathbf{x} &=\left(\mathbf{W}_{r} * \mathbf{q}_{r}-\mathbf{W}_{x} * \mathbf{q}_{x}-\mathbf{W}_{y} * \mathbf{q}_{y}-\mathbf{W}_{z} * \mathbf{q}_{z}\right) \\
&+\left(\mathbf{W}_{x} * \mathbf{q}_{r}+\mathbf{W}_{r} * \mathbf{q}_{x}-\mathbf{W}_{z} * \mathbf{q}_{y}+\mathbf{W}_{y} * \mathbf{q}_{z}\right) \mathbf{i} \\
&+\left(\mathbf{W}_{y} * \mathbf{q}_{r}+\mathbf{W}_{z} * \mathbf{q}_{x}+\mathbf{W}_{r} * \mathbf{q}_{y}-\mathbf{W}_{x} * \mathbf{q}_{z}\right) \mathbf{j} \\
&+\left(\mathbf{W}_{z} * \mathbf{q}_{r}-\mathbf{W}_{y} * \mathbf{q}_{x}+\mathbf{W}_{x} * \mathbf{q}_{y}+\mathbf{W}_{r} * \mathbf{q}_{z}\right) \mathbf{k} ,
\end{aligned}\label{14}
\end{equation}
\end{small}

\noindent and can be expressed in a matrix form:

\begin{equation}
\mathbf{W}_q \otimes \mathbf{x}=\left[\begin{array}{cccc}
\mathbf{W}_{r}&-\mathbf{W}_{x} & -\mathbf{W}_{y} & -\mathbf{W}_{z} \\
\mathbf{W}_{x}&\mathbf{W}_{r} & -\mathbf{W}_{z} & \mathbf{W}_{y} \\
\mathbf{W}_{y}&\mathbf{W}_{z} & \mathbf{W}_{r} & -\mathbf{W}_{x} \\
\mathbf{W}_{z}&-\mathbf{W}_{y} & \mathbf{W}_{x} & \mathbf{W}_{r}
\end{array}\right] \left[\begin{array}{c}
\mathbf{q}_{r} \\
\mathbf{q}_{x} \\
\mathbf{q}_{y} \\
\mathbf{q}_{z}
\end{array}\right].\label{15}
\end{equation}

We now show that the principle analysis of quaternion-valued correlation learning. A visual explanation of the quaternion-valued 2D convolution is shown in Figure~\ref{Figure3}, quaternion convolution allows the sharing of filters in channel dimensions, thus forcing each axis of the kernel to exploit the hidden internal relations in the quaternion space. Specifically, a quaternion kernel convolved against $\boldsymbol{Q}_{p}$ will perform a \textbf{double learning}: 1) the convolution operator learns outside (query) global relations among the elements of the query spatial dimension, 2) while the \textit{Hamilton product} accomplishes the inside (support) internal learning of the support subspace (quaternion space). This double learning model, therefore, is particularly suitable for correlation learning in FSS because it can exploit the information of two sub-dimensions (i.e., the support and query) and consider their interaction at the same time.

It is worth noticing the important difference in terms of the number of learning parameters between real and quaternion valued convolution. Denote $k$ as the number of kernels, $l$ as kernel size and $c$ as the number of input channels. In the case of a real-valued convolution layer with $k\;l \times l \times c$ kernels will have $kcl^2$ parameters, while to maintain equal $k$ and $c$ the quaternion equivalent has $\frac{k}{4}$ quaternion-valued kernels and $\frac{c}{4}$ quaternion-input channels. Therefore, the quaternion layers with $\frac{k}{4}\;l \times  l \times \frac{c}{4}$ has  $\frac{kcl^2}{16} \times 4 = \frac{kcl^2}{4}$ parameters: each kernel has 4 parameter variable elements, namely $\mathbf{W}^r$, $\mathbf{W}^x$, $\mathbf{W}^y$, $\mathbf{W}^z$. In other words, the degrees of freedom in Quaternion-valued convolution is only a quarter of those in its real-space counterpart. 

\subsubsection{Quaternion normalization}
The normalization\cite{RN69, RN70} is used to stabilize and speed up the training process of deep neural networks, which has been established as a very effective component in deep learning. The main idea behind normalization is to normalize inputs to have zero mean and unit variance along single or multiple dimensions. We notice that these formulations of normalization only work for real-values. Applying the above normalization to complex or hyper-complex numbers would be difficult since they can not simply translate and scale them such that their mean is 0 and their variance is 1. Therefore, we consider normalizing the quaternions using group normalization\cite{RN69}, which divides the channels into groups and computes within each group the mean and variance for normalization. 

However, normalizing within each group introduces problems—\textit{GN would not give equal variance in the multiple components of a quaternion, caused by independent variance calculations of each group}. To overcome this for complex numbers, we use the augmented covariance matrix in\cite{RN68} to recover the complete second-order statistics in the quaternion domain, which is defined as: 
\begin{equation}
\tilde{\mathbf{C}}_{\mathbf{q q}}=\mathrm{E}\left\{\tilde{\mathbf{q}} \tilde{\mathbf{q}}^{\mathrm{H}}\right\}=
\left[\begin{array}{cccc}
\mathbf{C}_{\mathbf{q}\mathbf{q}} & \mathbf{C}_{\mathbf{q}\mathbf{q}^{i}} & \mathbf{C}_{\mathbf{q}\mathbf{q}^{j}} & \mathbf{C}_{\mathbf{q}\mathbf{q}^{k}} \\
\mathbf{C}_{\mathbf{q}\mathbf{q}^{i}}^{\mathrm{H}} & \mathbf{C}_{\mathbf{q}^{i} \mathbf{q}^{i}} & \mathbf{C}_{\mathbf{q}^{i}\mathbf{q}^{j}} & \mathbf{C}_{\mathbf{q}^{i}\mathbf{q}^{k}} \\
\mathbf{C}_{\mathbf{q}\mathbf{q}^{j}}^{\mathrm{H}} & \mathbf{C}_{\mathbf{q}^{j}\mathbf{q}^{i}} & \mathbf{C}_{\mathbf{q}^{j}\mathbf{q}^{j}} & \mathbf{C}_{\mathbf{q}^{j}\mathbf{q}^{k}} \\
\mathbf{C}_{\mathbf{q}\mathbf{q}^{k}}^{\mathrm{H}} & \mathbf{C}_{\mathbf{q}^{k}\mathbf{q}^{i}} & \mathbf{C}_{\mathbf{q}^{k}\mathbf{q}^{j}} & \mathbf{C}_{\mathbf{q}^{k}\mathbf{q}^{k}} 
\end{array}\right],\label{16}
\end{equation}

\noindent where $(\cdot)^{\mathrm{H}}$ is the conjugate transpose operator, each $\mathbf{C}$ is the covariance between its two subscripts which represent the real, $i$, $j$, and $k$ components of $\tilde{\mathbf{q}} $ respectively. To make Equation~\ref{16} more feasible for practical applications, we further utilize the $\mathbb{Q}$-properness\cite{RN65} to simplify its computation. The $\mathbb{Q}$-properness implies that the quaternion vector $\mathbf{q}$ is not correlated with its vector involutions $\mathbf{q}^{i}$, $\mathbf{q}^{j}$, $\mathbf{q}^{k}$, i.e., $\mathbf{C}_{\mathbf{q}\mathbf{q}^{i}}=\mathbf{C}_{\mathbf{q}\mathbf{q}^{j}}=\mathbf{C}_{\mathbf{q}\mathbf{q}^{k}}=0$. 
Thus, considering a $\mathbb{Q}$-proper quaternion, the covariance in Equation~\ref{16} becomes: 
\begin{equation}
\begin{aligned} 
\tilde{\mathbf{C}}_{\mathbf{q q}}
&=\left[\begin{array}{cccc}
\mathbf{C}_{\mathbf{q}\mathbf{q}} & \mathbf{0}  & \mathbf{0}  & \mathbf{0}  \\
\mathbf{0}  & \mathbf{C}_{\mathbf{q}^{i} \mathbf{q}^{i}} & \mathbf{0}  & \mathbf{0}  \\
\mathbf{0}  & \mathbf{0}  & \mathbf{C}_{\mathbf{q}^{j}\mathbf{q}^{j}} & \mathbf{0}  \\
\mathbf{0}  & \mathbf{0}  & \mathbf{0}  & \mathbf{C}_{\mathbf{q}^{k}\mathbf{q}^{k}} 
\end{array}\right]  \\
&=\sum_{\delta\in \{r,x,y,z\}}\mathrm{E}\left\{\mathbf{q}_{\delta}^{2} \right\}\mathbf{I}
\end{aligned}.\label{17}
\end{equation}

Notwithstanding the above approach relies on the assumption that the input signal is $\mathbb{Q}$-proper, but it shows that the variance of a quaternion is obtained by the variance of its four components. Therefore, as an approximate of complete variance, we consider the average of the variance of each component as the quaternion variance and build the normalization as follows:
\begin{equation}
\mathrm{QN}(\mathbf{x})=(\frac{\mathbf{x}-\boldsymbol{\mu}_{q}}{\sqrt{\operatorname{var}\{\mathbf{x}\}+\mathbf{\epsilon}}})\mathbf{\gamma}+\mathbf{\beta}=(\frac{\mathbf{x}-\boldsymbol{\mu}_{q}}{\sqrt{\boldsymbol{\sigma}^{2}+\mathbf{\epsilon}}})\mathbf{\gamma}+\mathbf{\beta},\label{18}
\end{equation}
where $\beta$ is a shifting quaternion parameter, $\gamma$ is a scalar parameter, and both of them are learnable parameters. $\boldsymbol{\mu}_{q}$ is the quaternion input mean, which is a quaternion itself, and $\boldsymbol{\sigma}^{2}$ is real-valued variance. The  $\boldsymbol{\mu}_{q}$ and $\boldsymbol{\sigma}^{2}$ are defined as:
\begin{equation}
\begin{aligned}
\boldsymbol{\mu}_q &=\frac{1}{C} \sum_{c=1}^C \mathbf{q}_{r, c}+\mathbf{q}_{x, c} \mathbf{i}+\mathbf{q}_{y, c} \mathbf{j}+\mathbf{q}_{z, c} \mathbf{k} \\
&=\bar{\mathbf{q}}_{r}+\bar{\mathbf{q}}_{x} \mathbf{i}+\bar{\mathbf{q}}_{y} \mathbf{j}+\bar{\mathbf{q}}_{z} \mathbf{k}
\end{aligned},\label{19}
\end{equation}

\begin{equation}
\boldsymbol{\sigma}^{2}=\frac{1}{4C} \sum_{\delta\in \{r,x,y,z\}}\sum_{c=1}^{C}(\mathbf{q}_{\delta, c}-\bar{\mathbf{q}}_{\delta}) \otimes(\mathbf{q}_{\delta, c}-\bar{\mathbf{q}}_{\delta})^{*}.\label{20}
\end{equation}

To summarize, the quaternion convolutional block is defined as:
\begin{equation}
\widehat{\boldsymbol{Q}}_{p}=\mathcal{F}^{\text {qcl }}_{\text {p }}(\boldsymbol{Q}_{p})=\operatorname{ReLU}(\mathrm{QN}\left(\mathbf{W} \otimes \boldsymbol{Q}_{p}+\mathbf{b}\right)).\label{21}
\end{equation}

\begin{figure}[t]
 \includegraphics[width=8.8 cm]{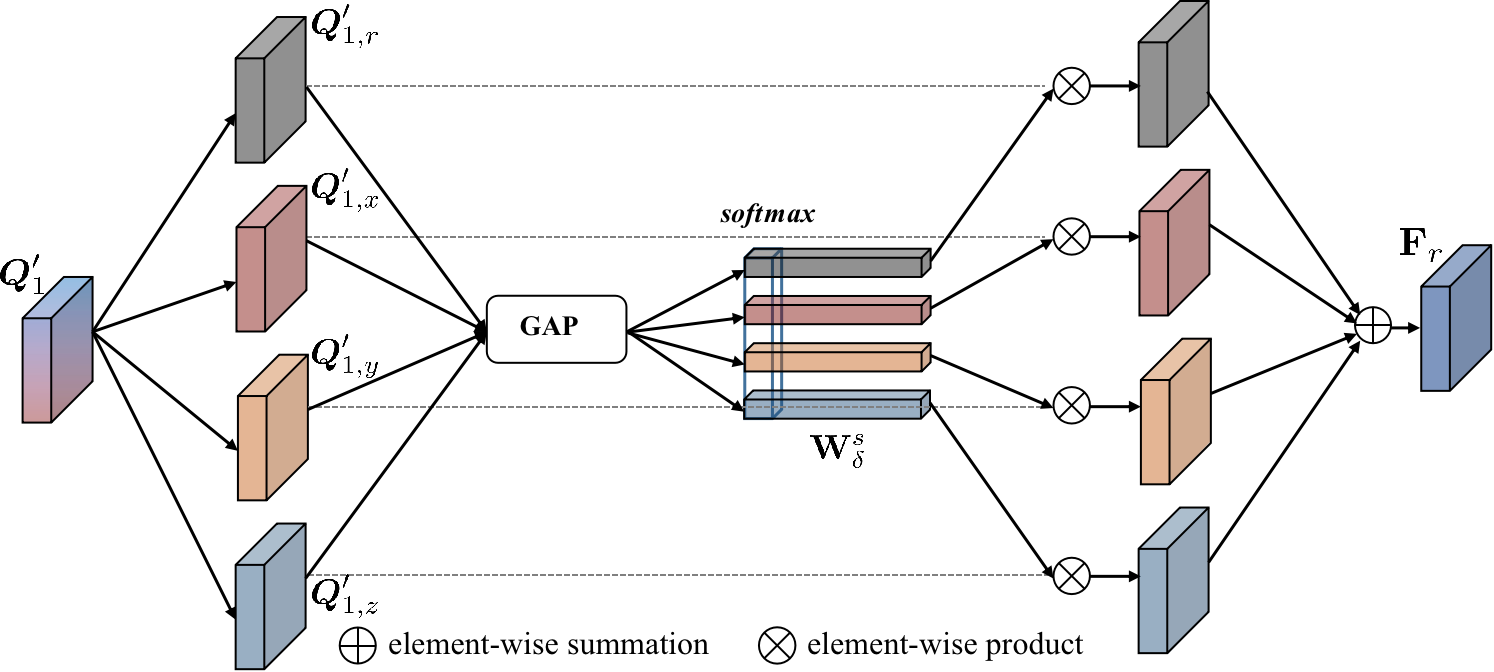}
\caption{ Visual illustration of the quaternion2real module. GAP means global average pooling. 
\label{Figure4}}
\end{figure}  

\subsubsection{Quaternion Aggregation}
In the quaternion aggregation module (QAM), the output of each $\widehat{\boldsymbol{Q}}_{p}$ is upsampled and element-wise summed with the next level $\widehat{\boldsymbol{Q}}_{p+1}$ with one degree of finer resolution. A quaternion convolution layer then processes this merged quaternion feature to propagate semantic information to finer branches in a coarse-to-fine fashion. By applying QAM to quaternion features at different spatial scales, finer quaternion feature maps can be guided using the rich semantic information of deeper-level features, which dramatically boosts the performance. The QAM is defined as:
\begin{equation}
\begin{aligned}
&\boldsymbol{Q}_{2}^{\prime}=\mathcal{F}^{\text {qcl }}(\widehat{\boldsymbol{Q}}_{2}+\mathrm{up}_{[\times 2]}(\widehat{\boldsymbol{Q}}_{3})) \\
&\boldsymbol{Q}_{1}^{\prime}=\mathcal{F}^{\text {qcl }}(\widehat{\boldsymbol{Q}}_{1}+\mathrm{up}_{[\times 2]}(\boldsymbol{Q}_{2}^{\prime}))
\end{aligned}.\label{22}
\end{equation}

\subsection{Episodic Readout Module}\label{4.7}

\begin{figure}[t]
 \includegraphics[width=8.4 cm]{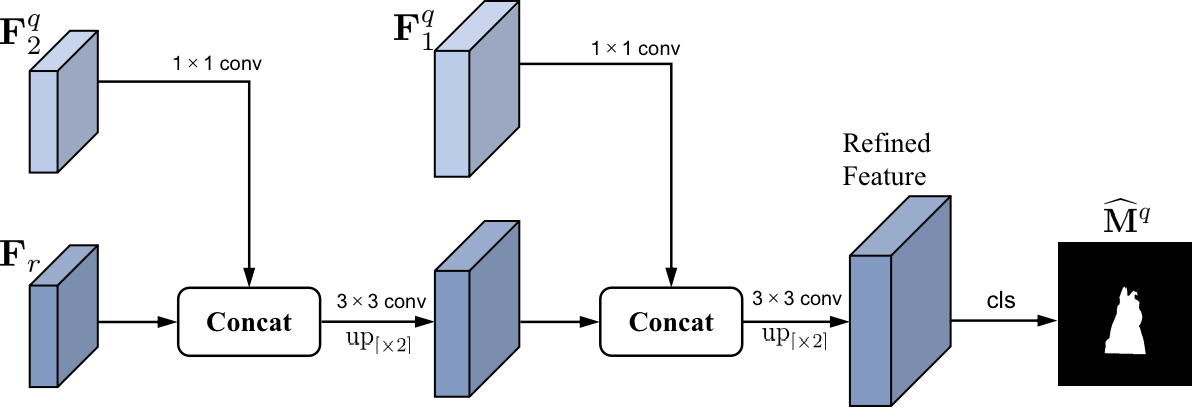}
\caption{Visual illustration of the real-valued convolutional decoder. $\mathrm{up}_{[\times 2]}$ is bilinear interpolation by a factor of 2 and $ \mathbf{F}_{i}^{q}$ is the low-level query feature extracted from \textit{block i}. 
\label{Figure5}}
\end{figure} 

\subsubsection{Transform Quaternion to Real}
Note that most tasks, such as semantic segmentation, require outputs composed of real numbers. However, the output of the quaternion correlation learning module $\boldsymbol{Q}_{1}^{\prime}$ consists of quaternions. Therefore, for FSS, we propose the quaternion2real module (Q2RM) to transform quaternion feature $\boldsymbol{Q}_{1}^{\prime}$ Into ordinary features (i.e., $\mathbb{H}^{H^p_1 \times W^p_1  \times D}\rightarrow \mathbb{R}^{H^p_1 \times W^p_1 \times D}$), in which each element is a real number. Specifically, as illustrated in Figure~\ref{Figure4}, we split $\boldsymbol{Q}_{1}^{\prime}$ into four components $\boldsymbol{Q}_{1,r}^{\prime}$, $\boldsymbol{Q}_{1,x}^{\prime}$, $\boldsymbol{Q}_{1,y}^{\prime}$, $\boldsymbol{Q}_{1,z}^{\prime}$ and embed the global information by simply using global average pooling (GAP) to generate channel-wise statistics. Then, the real-valued feature map $\mathbf{F}_{r}$ is obtained through the soft-attention weight $\mathbf{W}^{s}_{\delta}$ on four components $\delta\in \{r,x,y,z\}$:
\begin{equation}
\mathbf{F}_{r}=\sum_{\delta\in \{r,x,y,z\}}(\mathbf{W}^{s}_{\delta}\cdot\boldsymbol{Q}_{1,\delta}^{\prime}) \in \mathbb{R}^{H^1_q \times W^1_q \times D}.\label{23}
\end{equation}
$\mathbf{W}^{s}_{\delta}$ is defined as:
\begin{equation}
\mathbf{W}^{s}_{\delta}=\frac{e^{\mathcal{G}(\boldsymbol{Q}_{1,\delta}^{\prime}))}}{\sum_{\delta\in \{r,x,y,z\}}e^{\mathcal{G}(\boldsymbol{Q}_{1,\delta}^{\prime}))}},\label{24}
\end{equation}
where $\mathcal{G}(\cdot)$ is the global average pooling.

\subsubsection{Real-valued Convolutional Decoder}

In the work of \cite{RN24, RN6, RN9}, the features are bilinearly upsampled to the original image size, which may not successfully recover object segmentation details. Therefore, we propose a real-valued convolution decoder, as illustrated in Figure~\ref{Figure5}. The real-valued feature $\mathbf{F}_{r}$ is first concatenated with the corresponding low-level query features \cite{RN73} from the backbone (e.g., \textit{block1} and \textit{block2} in ResNet-50\cite{RN5} ) and then bilinear interpolation by a factor of 2. We apply a $1\times1$ convolution on the low-level query features to reduce the number of channels since the low-level features usually contain a large number of channels (e.g., 256 or 512) which may outweigh the importance of $\mathbf{F}_{r}$. After the concatenation, the merged features are refined by utilizing $3\times 3$ convolution and undergoes the above operations until the classification head which outputs the predicted mask $\mathbf{M}_{q}^{\prime}\in [0,1]^{H \times W \times 2}$. The process is defined as follows:
\begin{equation}
\mathbf{M}_{q}^{\prime}= \operatorname{Decoder}\left(\mathbf{F}_{r} , \mathcal{P}_{1}(\mathbf{F}_{1}^{q}), \mathcal{P}_{2}(\mathbf{F}_{2}^{q})\right),\label{25}
\end{equation}
where $\mathcal{P}_{i}(\cdot)$ linear projection. During testing, we take the maximum channel value at each pixel to obtain the predicted mask $\tilde{\mathbf{M}}_{q}\in\{0,1\}^{H \times W}$ for evaluation.

\subsection{Attention Mechanism for K-shot Segmentation}\label{4.8}
In order to efficiently merge semantic information in the K-shot setting, we use a prior attention mechanism to dynamically fuse the predictions generated by different support images. Specifically, we compute the cosine similarity of the last layer of query and K support features to obtain K correlation tensors $\{\mathbf{c}_{i}\left(\mathbf{x}^{\mathrm{q}},\mathbf{x}^{\mathrm{s}}\right)\}_{i=1}^{K}$, and then we take the maximum similarity among all support sub-dimension to generate the prior weight matrixes $\{\pmb{w}_{i}\}_{i=1}^{K} \in \mathbb{R}^{H^p_q \times W^p_q}$:
\begin{equation}
\pmb{w}_{i}=\max_{\mathbf{x}^{\mathrm{s}}} \mathbf{c}_{i}\left(\mathbf{x}^{\mathrm{q}}, \mathbf{x}^{\mathrm{s}}\right).\label{26}
\end{equation}

Since the prior weight matrix is obtained by calculating the highest correspondence from support spatial sub-dimension, they provide pixel-level prior information about which support sample is more important. Therefore, we multiply the predictions of each shot branch with the weight matrix normalized by the softmax function, i.e., $\tilde{\mathbf{M}}^{q}=\sum_{i=1}^{K}\operatorname{Softmax}(\pmb{w}_{i})\cdot \mathbf{M}_{q,i}^{\prime}$. If the final prediction is above threshold $\tau$ , we assign foreground pixels to it, otherwise assignb ackground:
\begin{equation}
\hat{\mathbf{M}}^{q}_{(x, y)}= \begin{cases}1 & \tilde{\mathbf{M}}^{q}_{(x,y)}>\tau \\ 0 & \text { otherwise }\end{cases}. \label{27}
\end{equation}
where $(x, y)$ denotes the spatial location.

\section{Experiments}
\subsection{Implementation Details}
\subsubsection{Datasets}
The experiments are conducted on two standard benchmark datasets of FSS: PASCAL-5$^{i}$\cite{RN74} and COCO-20$^{i}$\cite{RN75}. The PASCAL-5$^{i}$ dataset is obtained by combining PASCAL VOC 2012 with SBD\cite{RN77}, consisting of 20 object classes that are divided into 4 folds. The COCO-20$^{i}$\cite{RN75} is a more challenging dataset, and it consists of 80 classes divided into 4 folds. Following the training/validation strategy of previous work\cite{RN24,RN13,RN7,RN6,RN9} , we use three folds to build the training set, while the remaining fold is used to validate the model for cross-validation. During the evaluation, 1000 episodes (support-query pairs) from the test set are randomly selected to calculate the mean Intersection over Union (mIoU) and binary Intersection over Union (FBIoU) of all categories.

\begin{table*}
\centering
\caption{Performance of 1-shot and 5-shot semantic segmentation on the COCO-20$^{i}$. The best mean-IoUs are marked in bold.  \label{tab2}}
\begin{tabular}{c|c|ccccc|ccccc|c} 
\hline
\multirow{2}{*}{Method} & \multirow{2}{*}{Backbone} & \multicolumn{5}{c|}{mean-IoU (1-shot)}                                                                                     & \multicolumn{5}{c|}{mean-IoU (5-shot)}                                                                                        & \multirow{2}{*}{\begin{tabular}[c]{@{}c@{}}\#~learnable\\params\end{tabular}}  \\ 
\cline{3-12}
                        &                           & Fold-0                 & Fold-1                 & Fold-2                 & Fold-3                 & Mean                   & Fold-0                 & Fold-1                 & Fold-2                 & Fold-3                 & \multicolumn{1}{l|}{Mean} &                                                                                \\ 
\hline
PPNet \cite{RN7}                   & ResNet-50                 & 28.1                   & 30.8                   & 29.5                   & 27.7                   & 29.0                   & 39.0                   & 40.8                   & 37.1                   & 37.3                   & 38.5                      & 31.5M                                                                          \\
PMMs \cite{RN8}                    & ResNet-50                 & 29.3                   & 34.8                   & 27.1                   & 27.3                   & 29.6                   & 33.0                   & 40.6                   & 30.1                   & 33.3                   & 34.3                      & -                                                                              \\
RPMMs \cite{RN8}                   & ResNet-50                 & 29.5                   & 36.8                   & 28.9                   & 27.0                   & 30.6                   & 33.8                   & 42.0                   & 33.0                   & 33.3                   & 35.5                      & -                                                                              \\
PFENet \cite{RN6}                   & ResNet-50                 & 36.5                   & 38.6                   & 34.5                   & 33.8                   & 35.8                   & 36.5                   & 43.3                   & 37.8                   & 38.4                   & 39.0                      & 10.8M                                                                          \\
ASGNet \cite{RN9}                   & ResNet-50                 & -                      & -                      & -                      & -                      & 34.6                   & -                      & -                      & -                      & -                      & 42.5                      & 10.4M                                                                          \\
HSNet \cite{RN85}                   & ResNet-50                 & 36.3                   & 43.1                   & 38.7                   & 38.7                   & 39.2                   & 43.3                   & 51.3                   & 48.2                   & 45.0                   & 46.9                      & \textbf{2.6M}                                                                  \\ 
CAPL \cite{RN64}                    & ResNet-50                 & -                   & -                   & -                   & -                   & 39.8                   & -                   & -                   & -                   & -                   & 48.3                      & -                                                                  \\ 
\hline
Ours                    & ResNet-50                 & \textbf{39.8}          & \textbf{45.7}          & \textbf{42.5}          & \textbf{41.2}          & \textbf{42.3}          & \textbf{46.4}          & \textbf{53.0}          & \textbf{52.1}          & \textbf{48.6}          & \textbf{50.0}             & \textbf{2.6M}                                                 \\ 
\hline
FWB \cite{RN82}                      & ResNet-101                & 17.0                   & 18.0                   & 21.0                   & 28.9                   & 21.2                   & 19.1                   & 21.5                   & 23.9                   & 30.1                   & 23.7                      & 43.0M                                                                          \\
PFENet \cite{RN6}                   & ResNet-101                & 34.3                   & 33.0                   & 32.3                   & 30.1                   & 32.4                   & 38.5                   & 38.6                   & 38.2                   & 34.3                   & 37.4                      & 10.8M                                                                          \\
HFA \cite{RN84}                     & ResNet-101                & 28.6                   & 36.0                   & 30.1                   & 33.2                   & 32.0                   & 32.6                   & 42.1                   & 30.3                   & 36.1                   & 35.3                      & 36.5M                                                                          \\
SAGNN \cite{RN83}                    & ResNet-101                & 36.1                   & 41.0                   & 38.2                   & 33.5                   & 37.2                   & 40.9                   & 48.3                   & 42.6                   & 38.9                   & 42.7                      & -                                                                              \\
HSNet \cite{RN85}                   & ResNet-101                & 37.2                   & 44.1                   & 42.4                   & 41.3                   & 41.2                   & 45.9                   & 53.0                   & 51.8                   & 47.1                   & 49.5                      & \textbf{\textbf{2.6M}}                                                         \\ 
CAPL \cite{RN64}                    & ResNet-101                 & -                   & -                   & -                   & -                   & 42.8                   & -                   & -                   & -                   & -                   & 50.4                      & -                                                                  \\ 
\hline
Ours                    & ResNet-101                & \textbf{40.0}          & \textbf{45.5}          & \textbf{45.1}          & \textbf{43.6}          & \textbf{43.6}          & \textbf{46.9}          & \textbf{55.8}          & \textbf{53.6}          & \textbf{51.1}          & \textbf{51.9}             & \textbf{2.6M}                                                                  \\
\hline
\end{tabular}
\end{table*}

\begin{table*}
\centering
\caption{Performance of 1-shot and 5-shot semantic segmentation on the Pascal-5$^{i}$. The best mean-IoUs are marked in bold.}
\label{tab3}
\begin{tabular}{c|c|ccccc|ccccc|c} 
\hline
\multirow{2}{*}{Method} & \multirow{2}{*}{Backbone} & \multicolumn{5}{c|}{mean-IoU (1-shot)}                                                                                                                                                       & \multicolumn{5}{c|}{mean-IoU (5-shot)}                                                                                                                                                                   & \multirow{2}{*}{\begin{tabular}[c]{@{}c@{}}\# learnable\\params\end{tabular}}  \\ 
\cline{3-12}
                        &                           & Fold-0                                      & Fold-1                                      & Fold-2                                      & Fold-3                             & Mean          & Fold-0                                      & Fold-1                                      & Fold-2                                      & Fold-3                             & \multicolumn{1}{l|}{Mean} &                                                                                \\ 
\hline
AMP  \cite{RN81}                      & VGG-16                    & 41.9                                        & 50.2                                        & 46.7                                        & 34.7                               & 43.4          & 41.8                                        & 55.5                                        & 50.3                                        & 39.9                               & 46.9                      & 15.8M                                                                          \\
PANet \cite{RN22}                    & VGG-16                    & 42.3                                        & 58.0                                        & 51.1                                        & 41.2                               & 48.1          & 51.8                                        & 64.6                                        & 59.8                                        & 46.5                               & 55.7                      & 14.7M                                                                          \\
HSNet \cite{RN85}                    & VGG-16                    & 59.6                                        & 65.7                                        & \textbf{59.6}                                        & 54.0                               & 59.7          & 64.9                                        & \textbf{69.0}                                        & \textbf{64.1}                                        & 58.6                               & 64.1                      & \textbf{2.6M}                                                                  \\ 
\hline
Ours                    & VGG-16                    & \textbf{61.3}                                            & \textbf{66.8}                                             & 58.4                                            & \textbf{55.8}                                   & \textbf{60.6}              & \textbf{66.1}                                            & 68.5                                            & 63.2                                            & \textbf{58.8}                                   & \textbf{64.2}                          & \textbf{2.6M}                                                                               \\ 
\hline
CANet  \cite{RN24}                   & ResNet-50                 & 52.5                                        & 65.9                                        & 51.3                                        & 51.9                               & 55.4          & 55.5                                        & 67.8                                        & 51.9                                        & 53.2                               & 57.1                      & 19.0M                                                                          \\
PPNet \cite{RN7}                   & ResNet-50                 & 47.8                                        & 58.8                                        & 53.8                                        & 45.6                               & 51.5          & 58.4                                        & 67.8                                        & 64.9                                        & 56.7                               & 62.0                      & 31.5M                                                                          \\
PFENet \cite{RN6}                  & ResNet-50                 & 61.7                                        & 69.5                                        & 55.4                                        & 56.3                               & 60.8          & 63.1                                        & 70.7                                        & 55.8                                        & 57.9                               & 61.9                      & 10.8M                                                                          \\
ASGNet  \cite{RN9}                  & ResNet-50                 & 58.8                                        & 67.9                                        & 56.8                                        & 53.7                               & 59.3          & 63.7                                        & 70.6                                        & 64.2                                        & 57.4                               & 63.9                      & 10.4M                                                                          \\
SAGNN \cite{RN83}                    & ResNet-50                 & 64.7                                        & 69.6                                        & 57.0                                        & 57.2                               & 62.1          & 64.9                                        & 70.0                                        & 57.0                                        & 59.3                               & 62.8                      & -                                                                              \\
HSNet \cite{RN85}                    & ResNet-50                 & 64.3                                        & \textbf{70.7}                               & 60.3                                        & 60.5                               & 64.0          & 70.3                                        & 73.2                                        & \textbf{67.4}                               & \textbf{67.1}                      & \textbf{69.5}             & \textbf{2.6M}                                                                  \\
CAPL \cite{RN64}                    & ResNet-50                 & -                                           & -                                           & -                                           & -                                  & 62.2          & -                                           & -                                           & -                                           & -                                  & 67.1                      & -                                                                              \\ 
\hline
Ours                    & ResNet-50                 & \textbf{65.2}                               & 70.3                                        & \textbf{60.8}                              & \textbf{61.0}                      & \textbf{64.3} & \textbf{70.6}                               & \textbf{73.5}                               & 66.7                                        & \textbf{67.1}                      & \textbf{69.5}             & \textbf{2.6M}                                                                  \\ 
\hline
FWB \cite{RN82}                     & ResNet-101                & 51.3                                        & 64.5                                        & 56.7                                        & 52.2                               & 56.2          & 54.8                                        & 67.4                                        & 62.2                                        & 55.3                               & 59.9                      & 43.0M                                                                          \\
PFENet \cite{RN6}                  & ResNet-101                & 60.5                                        & 69.4                                        & 54.4                                        & 55.9                               & 60.1          & 62.8                                        & 70.4                                        & 54.9                                        & 57.6                               & 61.4                      & 10.8M                                                                          \\
ASGNet  \cite{RN9}                  & ResNet-101                & 59.8                                        & 67.4                                        & 55.6                                        & 54.4                               & 59.3          & 64.6                                        & 71.3                                        & 64.2                                        & 57.3                               & 64.4                      & 10.4M                                                                          \\
HSNet \cite{RN85}                    & ResNet-101                & 67.3                                        & 72.3                                        & 62.0                                        & 63.1                      & 66.2          & 71.8                                        & 74.4                                        & 67.0                                        & 68.3                      & 70.4                      & \textbf{2.6M}                                                                  \\
CAPL \cite{RN64}                     & ResNet-101                & -                                           & -                                           & -                                           & -                                  & 63.6          & -                                           & -                                           & -                                           & -                                  & 68.9                      & -                                                                              \\ 
\hline
Ours                    & ResNet-101                 & \textbf{67.9}                               & \textbf{72.5}                                        & \textbf{64.3}                              & \textbf{63.4}                      & \textbf{67.0} & \textbf{72.5}                               & \textbf{74.8}                               & \textbf{68.5}                                        & \textbf{68.9}                      & \textbf{71.2}             & \textbf{2.6M}                                                                  \\ 
\hline
\end{tabular}
\end{table*}

\begin{table}
\centering
\caption{Comparison of FB-IoU performance of 1-shot and 5-shot segmentation on the COCO-20$^{i}$. $\triangle$ means increment over 1-shot segmentation result.  \label{tab4}}
\begin{tabular}{c|c|c|c} 
\hline
Method & 1-shot & 5-shot & $\triangle$  \\ 
\hline
PANet \cite{RN22} & 59.2   & 63.5   & 4.3                                                           \\
PFENet \cite{RN6} & 58.6   & 61.9   & 3.3                                                           \\
DAN \cite{RN13}    & 62.3   & 63.9   & 1.6                                                           \\
ASGNet \cite{RN9} & 60.4   & 67.0   & \textbf{6.6}                                                           \\
SAGNN \cite{RN83}  & 60.9   & 63.4   & 2.5                                                          \\ 
HSNet \cite{RN85}  & 68.2   & 70.7   & 2.5                                                           \\ 
\hline
Ours   & \textbf{69.9}      & \textbf{73.5}      & 4.6                                                             \\
\hline
\end{tabular}
\end{table}

\subsubsection{Experimental Setting}
We use ResNet-50\cite{RN5} to conduct our main body experiments for a fair comparison with other methods. As in \cite{RN24,RN6}, all backbone networks are initialized with ImageNet\cite{RN78} pre-trained weights and are fixed during QCLNet training. The reason for doing so is to avoid them learning class-specific representations of the training data.  Other layers are initialized by the default setting of PyTorch and the quaternion parameters are initialized following the proposal of \cite{RN27}. It is worth mentioning that the number of quaternion feature maps is four times larger than real-valued, meaning 1 quaternion-valued feature map corresponds to 4 real-valued ones. To resolve the computational burden and preserve the representational ability of the model, the channel dimension $D$ is fixed as 64 in all of the experiments. For ResNet, the features after \textit{block2} are extracted to construct 3 pyramidal layers with different spatial scales. As such, by taking images with size $473\times 473$ as input for ResNet-50, we can get the feature map with spatial size $[60\times 60,30\times 30,15\times 15]$. In addition, we implement QCLNet using Pytorch\cite{RN79} on Nvidia 3060 GPUs. QCLNet is trained in an episode-based meta-learning fashion using the Adam optimization algorithm\cite{RN80} with a learning rate of $1e^{-3}$. The threshold $\tau$ in Eq.~\ref{27} for PASCAL-5$^{i}$ and COCO-20$^{i}$ is 0.5 and 0.6, respectively.

\subsubsection{Evaluation Metric}
We use the mean Intersection over Union (mIoU) and binary Intersection over Union (FB-IoU) as our evaluation metrics. For category $c$, IoU Is defined as $I o U_{k}=T P_{k} /\left(T P_{k}+F P_{k}+F N_{k}\right)$ where the $TP_{k}$, $FP_{k}$, $FN_{k}$ are the number of true positives, false positives, and false negatives in segmentation masks. The mIoU metric average the IoUs of all test categories in a fold. The formulation follows $mIoU=\frac{1}{C} \sum_{k=1}^{C} I o U_{i}$ where $C$ is the number of classes in each fold. And the FB-IoU calculates the mean of foreground and background IoUs regardless of the categories. As stated in \cite{RN24}, we mainly focus on mIoU since it  considers the differences of all classes so that the performance bias of scarce classes can be alleviated.

\subsection{Performance and Comparison}

\subsubsection{COCO-20$^{i}$}
The COCO-20$^i$ dataset is a very challenging dataset that contains many objects in a realistic scene image. We illustrate the mean-IoU in Table \ref{tab2}, from which it can be seen that QCLNet achieves state-of-the-art results with a competitive parameter size (2.6M) under both 1-shot and 5-shot settings. For instance, under the 1-shot setting, with a VGG16 backbone, we outperform the HSNet and CAPL methods by 3.1\% and 2.5\%. Under the 1-shot setting, with a ResNet50 backbone, QCLNet outperform the HSNet methods by 2.4\% and achieves the new state-of-the-art. In addition, we gain an impressive improvement of 3.1\% and 2.4\% in the 5-shot setting, which are significant margins for the challenging task. As such, the quaternion correlation learning in QCLNet indeed captures some inner benefits for boosting FSS performance, and we hope our model can shed light on future research in FSS.

Since most foreground classes only occupy a small spatial region of the whole image, the FB-IoU is biased toward the background class and causes it not convincing when evaluating performance. However, we also make comparisons of our model with other advanced approaches to COCO-20$^i$ and the numbers are competitive (see Table \ref{tab4}).

\subsubsection{PASCAL-5$^{i}$}
In Table \ref{tab3}, we compare QCLNet with the state-of-the-art methods on PASCAL-5$^{i}$. QCLNet outperforms state-of-the-art methods under both 1-shot and 5-shot settings. Specifically, in the 1-shot settings, our method outperforms the state-of-the-art by 0.3\% and 0.8\% with ResNet-50 and ResNet-101 respectively. And QCLNet performs significantly better than other methods by 0.8\% with the resnet101 backbone in the 5-shot setting. It is worth mentioning that PFENet and SAGNN used the additional training of the model for k-shot setting while QCLNet uses a simple k-shot fusion strategy to fuse the 5-shot results.

\begin{figure}
 \includegraphics[width=7.8 cm]{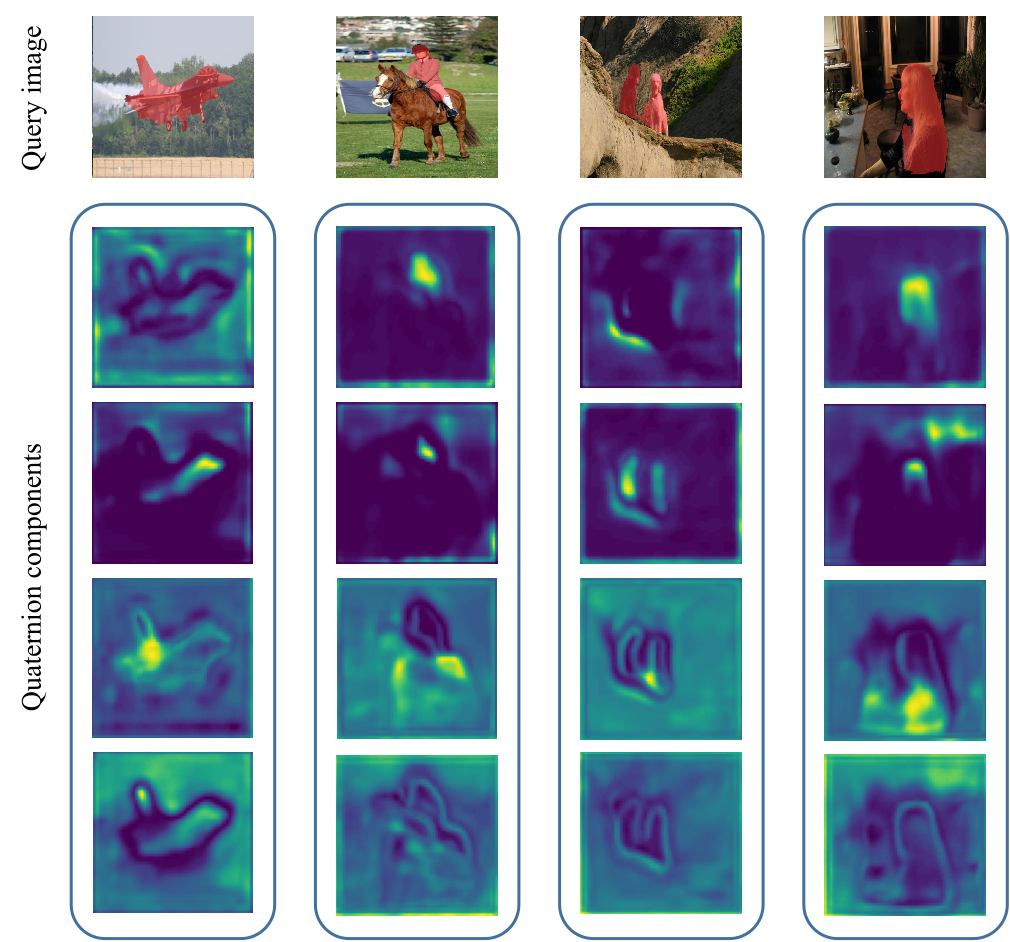}
\caption{Visualization of the four components of the quaternion feature. (Best viewed in color)
\label{Figure6}}
\end{figure} 

\begin{figure*}
\centering
\includegraphics[width=17.8 cm]{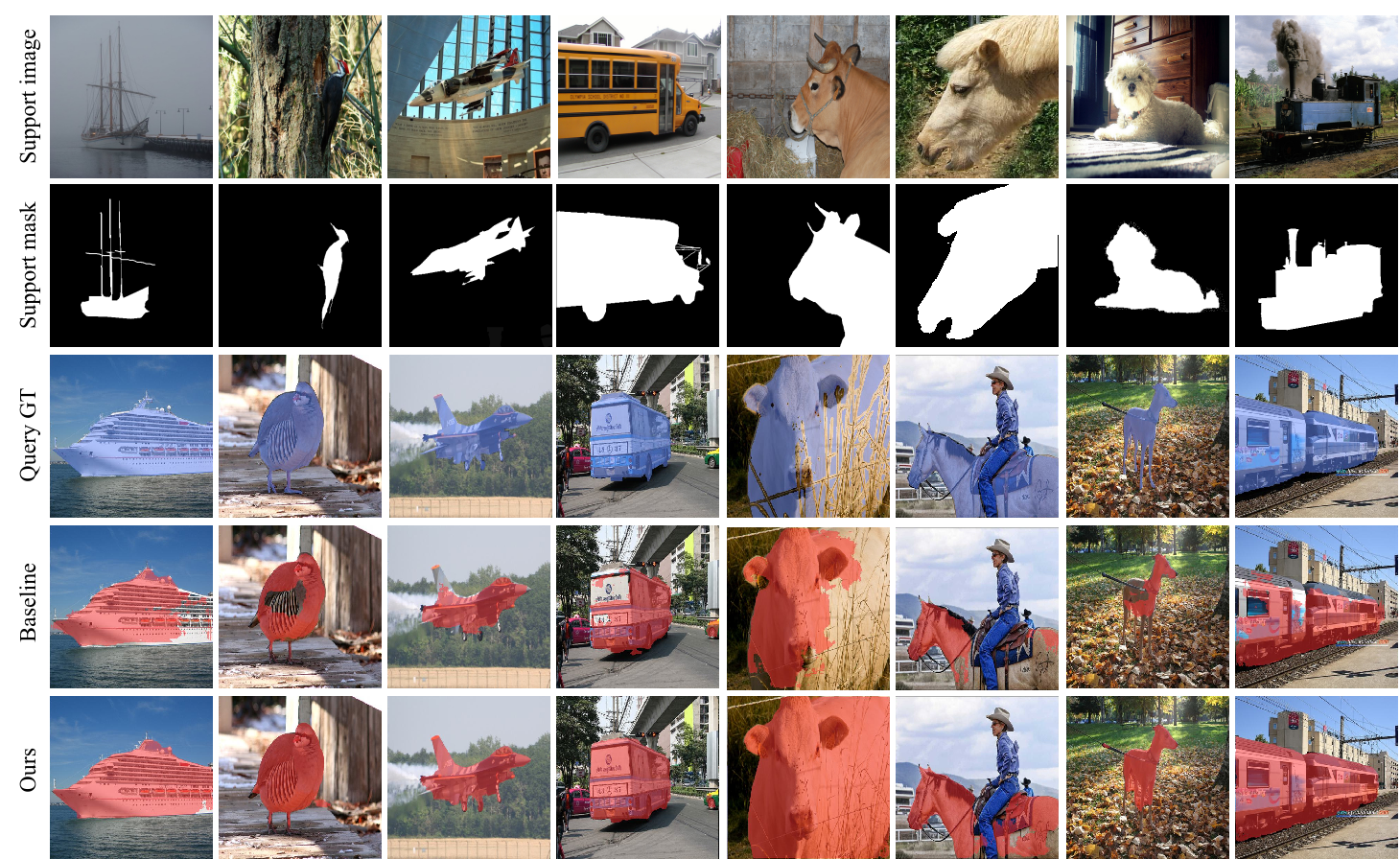}
\caption{Segmentation results of the proposed QCLNet and the baseline. From top to bottom: (a) support image, (b) support mask, (c) ground truth of query image, (d) predictions of baseline, (e) predictions of QCLNet.
\label{Figure7}}
\end{figure*}

\subsubsection{Segmentation Examples}
To better understand our proposed method, we show segmentation results during the meta-testing phase, as shown in Figure~\ref{Figure7}. It can be found in our method (4th row), the accurate and complete segmentation results of novel classes are apparently generated compared with the baseline method (3rd row), which verifies the effectiveness of quaternion-valued correlation learning. We further visualize the four components of quaternion features, Figure~\ref{Figure6}. Based on the interaction of quaternion-weight components and soft-attention Q2RM, QCLNet capture different key relations within the components of a quaternion.

\subsection{Ablation Studies}
In this section, we show an ablation analysis to inspect the effectiveness of our major contributions, justify the architectural choices we made and investigate whether QCL can effectively learn relations between support and query set while suppressing the degrees of freedom of model's parameter. Throughout this section, all the experiments are conducted with ResNet-50 in the 1-shot and 5-shot settings on PASCAL-5$^i$. Each ablation experiment is conducted under the same experimental setting for a fair comparison. We define the baseline as a model that replaces CAM with simple max pooling followed by standard 2D convolution, skips QCLM,  and disregards low-level query features. Then we evaluate the effectiveness of our methods by adding components progressively.

\subsubsection{Correlation aggregation}
CAM aims at aggregating sparse information in a high-dimensional correlation tensor before the correlation learning process. In Table \ref{tab5}, by simply introducing CAM to the baseline method, we improved the performance from 57.4\% and 63.1\% to 61.7\% and 67.8\%, demonstrating the necessity of aggregating support sub-dimension information. And CAM not only eases the computational burden caused by the high-dimensional properties of the correlation tensors, but also helps construct quaternion representations to facilitate subsequent correlation learning. 

\subsubsection{Quaternion Correlation Learning}
The CAM simply employs two real-valued 2D convolutions in two sub-dimensions, which lacks the ability to consider interactions of matching scores. Therefore, as shown in Table \ref{tab5}, with CAM followed by QCLM, we further improve the segmentation performance by 1.8\% and 0.7\%. This result shows that by encapsulating support spatial dimension as a quaternion, our method preserves intra-subspace relations and then enhances the hidden 
interactions between support and query. 

\begin{table}
\centering
\caption{Ablation study of the proposed approach on PASCAL-5$^{i}$. "CAM" denotes correlation aggregation module while "QCLM" denotes quaternion correlation learning module with quaternion normalization.  \label{tab5}}
\begin{tabular}{ll|cc} 
\hline
\multirow{2}{*}{} & \multirow{2}{*}{Components} & \multicolumn{2}{c}{mean-IoU}  \\ 
\cline{3-4}
                  & \multicolumn{1}{c|}{}                            & 1-shot & 5-shot               \\ 
\hline
\textbf{(I)}           & Baseline                                         & 57.4   & 63.1                 \\
\textbf{(II)}           & + CAM                                            & 61.7   & 67.8                 \\
\textbf{(III)}           & + QCLM                                           & 63.5   & 68.5                 \\
\textbf{(IV)}           & + Q2RM                                           & 64.0   & 68.8                 \\
\textbf{(V)}           & + Low-level feat                                 & \textbf{64.3}   & \textbf{69.5}                \\
\hline
\end{tabular}
\end{table}

\subsubsection{Effectiveness of Q2RM}
In Table \ref{tab5}, when using the Q2RM to transform quaternion features instead of simply averaging the quaternion components, the performance improvement is significant (0.5\% and 0.3\%), validating the importance of quaternion components statistics because they correspond to different support subspaces. 

\subsubsection{Effectiveness of Low-level Feature}
We also demonstrate the effectiveness of the low-level feature. The results of Table \ref{tab5} show that adding low-level features in the decoder can help to find accurate correspondences, which in turn yields better segmentation performance (0.3\% and 0.7\%). We consider this improvement is that the low-level feature contains fine object segmentation detail, which can resolve the ambiguities in the correlation map and expedite the learning process.

\begin{table}
\centering
\caption{Performance under different correlation aggregate strategies. "Top K" denotes aggregates correlation by storing top k promising matching scores, "Sep 4D conv" denote Separable 4D convolution. \label{tab6}}
\begin{tabular}{l|cc} 
\hline
\multirow{2}{*}{Different aggregators} & \multicolumn{2}{c}{mean-IoU}                 \\ 
\cline{2-3}
                                       & 1-shot                             & 5-shot  \\ 
\hline
\multicolumn{1}{c|}{TopK}              &62.6 & 66.9    \\
\multicolumn{1}{c|}{Sep 4D conv}       & \textbf{64.3}  & \textbf{69.5}    \\
\hline
\end{tabular}
\end{table}

\subsubsection{Ablation Study on CAM}
As mentioned in Section \ref{4.4}, we propose to aggregate the sparse information by applying separable 4D convolution. Compared with simply storing Top K promising matching scores, the proposed module achieves a sizable gain (see Table \ref{tab6}) under 1-shot and 5-shot settings. We attribute this phenomenon to the different utilization of neighborhood information by the two methods. Specifically, one tends to use convolution to gradually aggregate sparse information in the the correlation tensor, while the other tends to keep only the top K matches into a sparse correlation tensor, which is challenging to get enough correlation statistics. 

\begin{table*}
\centering
\caption{Performance under three different 2D convonlution kernels. “Gconv” denotes the group convolution, "Sconv" denotes the standard convolution and "Qconv" denotes the quaternion-valued convolution. \label{tab7}}
\begin{tabular}{c|ccccc|ccccc|c} 
\hline
\multirow{2}{*}{Kernel type} & \multicolumn{5}{c|}{mean-IoU (1-shot)}                                        & \multicolumn{5}{c|}{mean-IoU (5-shot)}                                                    & \multirow{2}{*}{\begin{tabular}[c]{@{}c@{}}\# learnable\\params\end{tabular}}  \\ 
\cline{2-11}
                             & Fold-0        & Fold-1        & Fold-2        & Fold-3        & Mean          & Fold-0        & Fold-1        & Fold-2        & Fold-3        & \multicolumn{1}{l|}{Mean} &                                                                                \\ 
\hline
Gconv kernel                 & 63.9          & 69.0          & \textbf{60.9} & 55.9          & 62.4          & 69.4          & 72.3          & 65.4          & 60.5          & 66.9                      & \textbf{2.6M}                                                                  \\
Sconv kernel                 & 64.4          & \textbf{70.6} & 60.3          & \textbf{61.1} & 64.1          & 69.9          & 73.1          & 65.6          & \textbf{67.1} & 68.9                      & 9.2M                                                                           \\
Qconv kernel                 & \textbf{65.2} & 70.3          & 60.8          & 61.0          & \textbf{64.3} & \textbf{70.6} & \textbf{73.5} & \textbf{66.7} & \textbf{67.1} & \textbf{69.5}             & \textbf{2.6M}                                                                  \\
\hline
\end{tabular}
\end{table*}

\begin{table}
\centering
\caption{Performance under different normalization strategies in QCL module. "BN" denotes batch normalization, "GN" denotes group normalization and "QN" denotes the quaternion normalization.}
\label{tab8}
\begin{tabular}{c|cc} 
\hline
\multicolumn{1}{l|}{\multirow{2}{*}{Different normalization}} & \multicolumn{2}{c}{mean-IoU}   \\ 
\cline{2-3}
\multicolumn{1}{l|}{}                                         & 1-shot        & 5-shot         \\ 
\hline
BN                                                            & 63.7          & 67.6           \\
GN                                                            & 64.2          & 69.0           \\
QN                                                            & \textbf{64.3} & \textbf{69.5}  \\
\hline
\end{tabular}
\end{table}

\subsubsection{Ablation Study on Different 2D Kernels}
We provide an ablation study on different 2D kernels to justify the use of our proposed quaternion-valued kernel for correlation learning. In specific,  we replace the proposed quaternion-valued kernel with the group convolution kernel and the standard 2D convolution kernel, and leaving all the other components for a fair comparison. Table \ref{tab7} summarizes the results. The learnable params are the parameters of the entire model after replacing the kernel. 

As shown in Table \ref{tab7}, while both the standard 2D kernel and our quaternion-valued kernel interact with different components in the channel dimension, the high dimensional space of hypercorrelation results in standard 2D convolution with four larger parameters (9.2M vs. 2.6M) and worse performance than ours. The performance gap indicates that our approach is more balanced mining the semantic relations of two subspaces and significantly improves model parameters. In addition, to demonstrate the effectiveness of the weight sharing strategy in Hamilton product, we also added a grouped convolution kernel in Table 8 for comparison, which gives a separate weight for each components in the quaternion feature and has the same parameter size (2.6M vs. 2.6M) as the quaternion convolution. The group convolution is defined as:

\begin{equation}
\begin{aligned}
\mathbf{W}_g \otimes \mathbf{x} & = \left(\mathbf{W}_{r} * \mathbf{q}_{r}\right)+\left(\mathbf{W}_{x} * \mathbf{q}_{x}\right)\mathbf{i} \\
& +\left(\mathbf{W}_{y} * \mathbf{q}_{y}\right)\mathbf{j}+\left(\mathbf{W}_{z} * \mathbf{q}_{z}\right)\mathbf{k}.
\end{aligned}\label{28}
\end{equation}

In Table \ref{tab7}, compared to group convolution which simply performs independent convolution performing independent convolution of the four components of a quaternion, the QCNN using Hamilton product improves the segmentation performance by 1.9\% and 2.6\%. This indicates that the weight sharing strategy of Hamilton product significantly maintains the important structural information of the quaternion space and fully exploit the internal latent interrelationship among the four components of quaternion feature.

\begin{figure}[t]
 \includegraphics[width=9.0 cm]{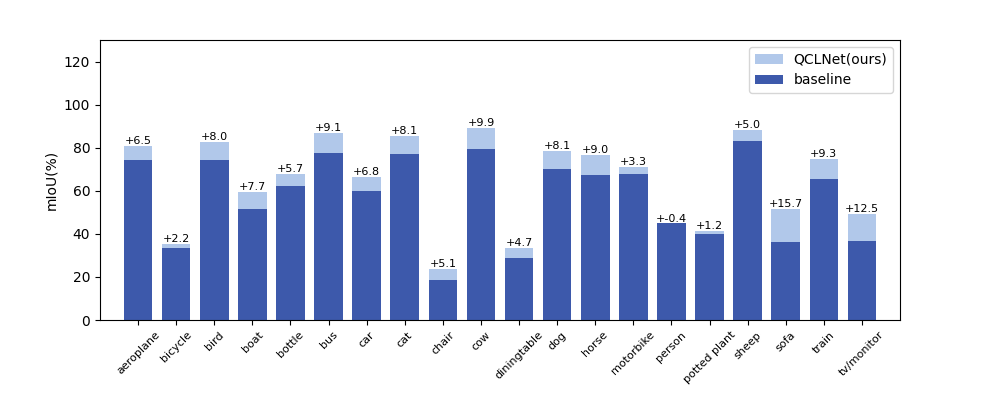}
\caption{Per-class performance gains on PASCAL-5$^{i}$ dataset. Our proposed QCLNet achieves significant improements against the baseline.
\label{Figure8}}
\end{figure} 

\begin{table}
\centering
\caption{Comparison of different 5-shot solutions. Our attention mechanism performs the best and brings the most increment over 1-shot baseline. }
\label{tab9}
\begin{tabular}{c|cc} 
\hline
Method           & mean-IoU & Increment  \\ 
\hline
1-shot baseline  & 64.3     & 0          \\ 
\hline
Mask-avg         & 69.3     & 5.0        \\
Attention (ours) &\textbf{69.5}    &\textbf{5.2}        \\
\hline
\end{tabular}
\end{table}

\subsubsection{Ablation Study on QN}
The normalization of quaternion features is an important component, which affects the stability of the complex-valued neural network training process. As shown in Table \ref{tab8}, with the replacement of QN with BN and GN in the QCLM, we improve the segmentation performance by 1.9 \% and 0.5\% in the 5-shot setting. This shows that in the quaternion space , the variance of each component should be calculated jointly and kept consistent. It is for this reason that the QN can effectively stabilize the training process of quaternion neural networks.

\subsubsection{Ablation Study on K-shot Fusion Schemes}
In the k-shot setting, we compare our attention mechanism to existing k-shot fusion scheme. We report the results of QCLNet with different fusion solutions in Table \ref{tab9}. Our attention mechanism performs the best and brings the most increment over a 1-shot baseline. This indicates that the training-free prior information can be more effective in fusing information from different support examples, which can effectively indicate which support sample is more important in the FSS setting with large intra-class variance.

\subsection{Extensions}
\subsubsection{Category-Wise Performance}

In Figure~\ref{Figure8}, we compare the category-wise segmentation performance on PASCAL-5$^{i}$. Categories such as cow, tv/monitor, train, and sofa achieved the largest performance gains. These categories can be largely affected by object views and pose (i.e., the query and support images may be quite different). This result shows that our proposed QCLNet has the potential to exploit the latent relationship between query and support images.

\begin{table}
\centering
\caption{Mean and Std. of five test results (class mIoU) on PASCAL-5$^{i}$ . Each row shows five test results with the values of mean and standard deviation (Std.).}
\label{tab10}
\begin{tabular}{c|ccccc|cc} 
\hline
Fold & Exp-1                              & Exp-2                              & Exp-3                              & Exp-4                              & Exp-5                              & Mean                               & Std.                                 \\ 
\hline
0    & {65.2} & {65.4} &{64.2} & {65.5} & {64.5} & {65.0} & {0.579}  \\
1    & {70.3} & {71.2} & {69.6} & {69.9} & {70.8} & {70.4} & {0.652}  \\
2    & {60.8} & {60.5} & {60.9} & {61.6} & {60.3} & {60.8} & {0.497}  \\
3    & {61.0} & {61.0} & {61.8} & {60.6} & {60.1} & {60.9} & {0.624}  \\
\hline
\end{tabular}
\end{table}

\subsubsection{Result Stability}
To demonstrate the stability and robustness of our model, we conduct multiple experiments on our PASCAL-trained QCLNet with different support samples. As seen in Table \ref{tab10}, the values of standard deviation are lower than 0.7, which shows that QCLNet is insensitive to support samples and exhibits strong stability.

\section{Conclusion}
We propose a novel Quaternion-valued Correlation Learning Network (QCLNet) for FSS to alleviate the class bias problem and explore precise support-query dense relations by considering the support sub-dimension in the quaternion space. Our QCLNet is formulated as a  hyper-complex valued network, which exploits the properties of quaternion algebra and captures internal latent relations between query and support set. It also shares the quaternion-weight components during the Hamilton product, which greatly suppresses the degrees of freedom of the model's parameters. Experiments have been conducted On the PASCAL-5$^{i}$ and COCO-20$^{i}$ datasets, and the results demonstrate that our method improves the performance of FSS, and is striking in contrast with the prior approaches. As a systematic yet theoretically based method for dense correlation learning, our QCLNet provides a novel perspective on the few-shot learning problem.

\section*{Acknowledgments}
This work was supported in part by Science and technology research in key areas in Foshan under Grant 2020001006832, the Key-Area Research and Development Program of Guangdong Province under Grant 2018B010109007 and 2019B010153002, the Science and technology projects of Guangzhou under Grant 202007040006, and the Guangdong Provincial Key Laboratory of Cyber-Physical System under Grant 2020B1212060069.


\begin{thebibliography}{10}

\bibitem{RN2}
O. Ronneberger, P. Fischer, and T. Brox, “U-net: Convolutional networks for biomedical image segmentation,” in \textit{Proc. Int. Conf. Med. Image Comput. Comput.-Assisted Intervention}. Springer, 2015, pp. 234–241.

\bibitem{RN3}
H. Zhao, J. Shi, X. Qi, X. Wang, and J. Jia, “Pyramid scene parsing network,” in \textit{Proc. IEEE Conf. Comput. Vis. Pattern Recognit.}, 2017, pp. 2881–2890.

\bibitem{RN33}
S. Shao, L. Xing, R. Xu, W. Liu, Y.-J. Wang, and B.-D. Liu, “Mdfm: Multi-decision fusing model for few-shot learning,” \textit{IEEE Trans. Circuits Syst. Video Technol.}, 2021.

\bibitem{RN34}
W. Jiang, K. Huang, J. Geng, and X. Deng, “Multi-scale metric learning for few-shot learning,” \textit{IEEE Trans. Circuits Syst. Video Technol.}, vol. 31, no. 3, pp. 1091–1102, 2020.

\bibitem{RN21}
N. Dong and E. P. Xing, “Few-shot semantic segmentation with prototype learning,” in \textit{Proc. Brit. Mach. Vis. Conf.}, vol. 3, 2018.

\bibitem{RN24}
C. Zhang, G. Lin, F. Liu, R. Yao, and C. Shen, “Canet: Class-agnostic segmentation networks with iterative refinement and attentive few-shot learning,” in \textit{Proc. IEEE Conf. Comput. Vis. Pattern Recognit.}, 2019, pp. 5217–5226.

\bibitem{RN9}
G. Li, V. Jampani, L. Sevilla-Lara, D. Sun, J. Kim, and J. Kim, “Adaptive prototype learning and allocation for few-shot segmentation,” in \textit{Proc. IEEE Conf. Comput. Vis. Pattern Recognit.}, 2021, pp. 8334–8343.

\bibitem{RN8}
B. Yang, C. Liu, B. Li, J. Jiao, and Q. Ye, “Prototype mixture models for few-shot semantic segmentation,” in \textit{Proc. Eur. Conf. Comput. Vis}. Springer, 2020, pp. 763–778.

\bibitem{RN22}
K. Wang, J. H. Liew, Y. Zou, D. Zhou, and J. Feng, “Panet: Few-shot image semantic segmentation with prototype alignment,” in \textit{Proc. IEEE Int. Conf. Comput. Vis.}, 2019, pp. 9197–9206.

\bibitem{RN6}
Z. Tian, H. Zhao, M. Shu, Z. Yang, R. Li, and J. Jia, “Prior guided feature enrichment network for few-shot segmentation,” \textit{IEEE Trans. Pattern Anal. Mach. Intell.}, no. 01, pp. 1–1, 2020.

\bibitem{RN38}
J. Min and M. Cho, “Convolutional hough matching networks,” in \textit{Proc. IEEE Conf. Comput. Vis. Pattern Recognit.}, 2021, pp. 2940–2950.

\bibitem{RN85}
J. Min, D. Kang, and M. Cho, “Hypercorrelation squeeze for few-shot segmentation,” in \textit{Proc. IEEE Int. Conf. Comput. Vis.}, 2021, pp. 6941– 6952.

\bibitem{RN19}
I. Rocco, M. Cimpoi, R. Arandjelović, A. Torii, T. Pajdla, and J. Sivic,  “Neighbourhood consensus networks,” \textit{Proc. Adv. Neural Inf. Process. Syst.}, vol. 31, 2018.

\bibitem{RN42}
I. Rocco, R. Arandjelović, and J. Sivic, “Efficient neighbourhood consensus networks via submanifold sparse convolutions,” in \textit{Proc. Eur. Conf. Comput. Vis}. Springer, 2020, pp. 605–621.

\bibitem{RN43}
V. Badrinarayanan, A. Kendall, and R. Cipolla, “Segnet: A deep convolutional encoder-decoder architecture for image segmentation,” \textit{IEEE Trans. Pattern Anal. Mach. Intell.}, vol. 39, no. 12, pp. 2481–2495, 2017.

\bibitem{RN44}
K. Sun, B. Xiao, D. Liu, and J. Wang, “Deep high-resolution representation learning for human pose estimation,” in \textit{Proc. IEEE Conf. Comput. Vis. Pattern Recognit.}, 2019, pp. 5693–5703.

\bibitem{RN46}
H. Noh, S. Hong, and B. Han, “Learning deconvolution network for semantic segmentation,” in \textit{Proc. IEEE Int. Conf. Comput. Vis.}, 2015, pp. 1520–1528.

\bibitem{RN48}
L.-C. Chen, G. Papandreou, I. Kokkinos, K. Murphy, and A. L. Yuille, “Deeplab: Semantic image segmentation with deep convolutional nets, atrous convolution, and fully connected crfs,” \textit{IEEE Trans. Pattern Anal. Mach. Intell.}, vol. 40, no. 4, pp. 834–848, 2017.

\bibitem{RN53}
J. Dai, H. Qi, Y. Xiong, Y. Li, G. Zhang, H. Hu, and Y. Wei, “Deformable convolutional networks,” in \textit{Proc. IEEE Int. Conf. Comput. Vis.}, 2017, pp. 764–773.

\bibitem{RN59}
Z. Zhu, M. Xu, S. Bai, T. Huang, and X. Bai, “Asymmetric non-local neural networks for semantic segmentation,” in \textit{Proc. IEEE Int. Conf. Comput. Vis.}, 2019, pp. 593–602.

\bibitem{RN20}
A. Shaban, S. Bansal, Z. Liu, I. Essa, and B. Boots, “One-shot learning for semantic segmentation,” in \textit{Proc. Brit. Mach. Vis. Conf.}, 2017.

\bibitem{RN7}
Y. Liu, X. Zhang, S. Zhang, and X. He, “Part-aware prototype network for few-shot semantic segmentation,” in \textit{Proc. Eur. Conf. Comput. Vis}. Springer, 2020, pp. 142–158.

\bibitem{RN64}
Z. Tian, X. Lai, L. Jiang, S. Liu, M. Shu, H. Zhao, and J. Jia, “Generalized few-shot semantic segmentation,” in \textit{Proc. IEEE Conf. Comput. Vis. Pattern Recognit.}, 2022, pp. 11 563–11 572.

\bibitem{RN40}
J. Min, J. Lee, J. Ponce, and M. Cho, “Learning to compose hypercolumns for visual correspondence,” in \textit{Proc. Eur. Conf. Comput. Vis}. Springer, 2020, pp. 346–363.

\bibitem{RN32}
S. Huang, Q. Wang, S. Zhang, S. Yan, and X. He, “Dynamic context correspondence network for semantic alignment,” in \textit{Proc. IEEE Int. Conf. Comput. Vis.}, 2019, pp. 2010–2019.

\bibitem{RN60}
S. Gai and X. Huang, “Reduced biquaternion convolutional neural network for color image processing,” \textit{IEEE Trans. Circuits Syst. Video Technol.}, vol. 32, no. 3, pp. 1061–1075, 2021.

\bibitem{RN61}
S.-C. Pei and C.-M. Cheng, “Color image processing by using binary quaternion-moment-preserving thresholding technique,” \textit{IEEE Trans. Image Process.}, vol. 8, no. 5, pp. 614–628, 1999.

\bibitem{RN62}
C. Papaioannidis and I. Pitas, “3d object pose estimation using multi-objective quaternion learning,” \textit{IEEE Trans. Circuits Syst. Video Technol.}, vol. 30, no. 8, pp. 2683–2693, 2019.

\bibitem{RN63}
A. Hirose and S. Yoshida, “Generalization characteristics of complex-valued feedforward neural networks in relation to signal coherence,” \textit{IEEE Trans. Neural Networks Learn. Syst.}, vol. 23, no. 4, pp. 541–551, 2012.

\bibitem{RN26}
C. J. Gaudet and A. S. Maida, “Deep quaternion networks,” in \textit{Int. Jt. Conf. Neural Networks}. IEEE, 2018, pp. 1–8.

\bibitem{RN27}
T. Parcollet, M. Morchid, and G. Linarès, “Quaternion convolutional neural networks for heterogeneous image processing,” in \textit{Proc. IEEE Int. Conf. Acoust. Speech Signal Process}. IEEE, 2019, pp. 8514–8518.

\bibitem{RN28}
Y. Tay, A. Zhang, A. T. Luu, J. Rao, S. Zhang, S. Wang, J. Fu, and S. C. Hui, “Lightweight and efficient neural natural language processing with quaternion networks,” in \textit{Proc. Annu. Meet. Assoc. Comput Linguist.}, 2019, pp. 1494–1503.

\bibitem{RN29}
X. Zhu, Y. Xu, H. Xu, and C. Chen, “Quaternion convolutional neural networks,” in \textit{Proc. Eur. Conf. Comput. Vis.}, 2018, pp. 631–647.

\bibitem{RN18}
G. Yang and D. Ramanan, “Volumetric correspondence networks for optical flow,” \textit{Proc. Adv. Neural Inf. Process. Syst.}, vol. 32, pp. 794– 805, 2019.

\bibitem{RN69}
Y. Wu and K. He, “Group normalization,” in \textit{Proc. Eur. Conf. Comput. Vis.}, 2018, pp. 3–19.

\bibitem{RN70}
S. Ioffe and C. Szegedy, “Batch normalization: Accelerating deep network training by reducing internal covariate shift,” in \textit{Proc. Int. Conf. Mach. Learn.} PMLR, 2015, pp. 448–456.

\bibitem{RN68}
E. Grassucci, E. Cicero, and D. Comminiello, \textit{Quaternion generative adversarial networks}. Springer, 2022, pp. 57–86.

\bibitem{RN65}
C. C. Took and D. P. Mandic, “Augmented second-order statistics of quaternion random signals,” \textit{Signal Process.}, vol. 91, no. 2, pp. 214– 224, 2011.

\bibitem{RN73}
B. Hariharan, P. Arbeláez, R. Girshick, and J. Malik, “Hypercolumns for object segmentation and fine-grained localization,” in \textit{Proc. IEEE Conf. Comput. Vis. Pattern Recognit.}, 2015, pp. 447–456.

\bibitem{RN5}
K. He, X. Zhang, S. Ren, and J. Sun, “Deep residual learning for image recognition,” in \textit{Proc. IEEE Conf. Comput. Vis. Pattern Recognit.}, 2016, pp. 770–778.

\bibitem{RN74}
M. Everingham, L. Van Gool, C. K. Williams, J. Winn, and A. Zisserman, “The pascal visual object classes (voc) challenge,” \textit{Int. J. Comput. Vis.}, vol. 88, no. 2, pp. 303–338, 2010.

\bibitem{RN75}
T.-Y. Lin, M. Maire, S. Belongie, J. Hays, P. Perona, D. Ramanan, P. Dollar, and C. L. Zitnick, “Microsoft coco: Common objects in context,” in \textit{Proc. Eur. Conf. Comput. Vis.} Springer, 2014, pp. 740–755.

\bibitem{RN77}
B. Hariharan, P. Arbeláez, R. Girshick, and J. Malik, “Simultaneous detection and segmentation,” in \textit{Proc. Eur. Conf. Comput. Vis.} Springer, 2014, pp. 297–312.

\bibitem{RN13}
H. Wang, X. Zhang, Y. Hu, Y. Yang, X. Cao, and X. Zhen, “Few-shot semantic segmentation with democratic attention networks,” in \textit{Proc. Eur. Conf. Comput. Vis.} Springer, 2020, pp. 730–746.

\bibitem{RN82}
K. Nguyen and S. Todorovic, “Feature weighting and boosting for few-shot segmentation,” in \textit{Proc. IEEE Int. Conf. Comput. Vis.}, 2019, pp. 622–631.

\bibitem{RN84}
B. Liu, J. Jiao, and Q. Ye, “Harmonic feature activation for few-shot semantic segmentation,” \textit{IEEE Trans. Image Process.}, vol. 30, pp. 3142– 3153, 2021.

\bibitem{RN83}
G.-S. Xie, J. Liu, H. Xiong, and L. Shao, “Scale-aware graph neural network for few-shot semantic segmentation,” in \textit{Proc. IEEE Conf. Comput. Vis. Pattern Recognit.}, 2021, pp. 5475–5484.

\bibitem{RN81}
M. Siam, B. N. Oreshkin, and M. Jagersand, “Amp: Adaptive masked proxies for few-shot segmentation,” in \textit{Proc. IEEE Int. Conf. Comput. Vis.}, 2019, pp. 5249–5258.

\bibitem{RN78}
J. Deng, W. Dong, R. Socher, L.-J. Li, K. Li, and L. Fei-Fei, “Imagenet: A large-scale hierarchical image database,” in \textit{Proc. IEEE Conf. Comput. Vis. Pattern Recognit.} IEEE, 2009, pp. 248–255.

\bibitem{RN79}
A. Paszke, S. Gross, F. Massa, A. Lerer, J. Bradbury, G. Chanan, T. Killeen, Z. Lin, N. Gimelshein, and L. Antiga, “Pytorch: An imperative style, high-performance deep learning library,” \textit{Proc. Adv. Neural Inf. Process. Syst.}, vol. 32, 2019.

\bibitem{RN80}
D. P. Kingma and J. Ba, “Adam: A method for stochastic optimization,” in \textit{Proc. Int. Conf. Learn. Represent.}, 2015.


\end{thebibliography}

\end{document}